\documentclass[10pt,twocolumn,letterpaper]{article}

\usepackage{cvpr}
\usepackage{times}
\usepackage{epsfig}
\usepackage{graphicx}
\usepackage{amsmath}
\usepackage{amssymb}
\usepackage{subcaption}
\usepackage[numbers, compress]{natbib}
\usepackage{booktabs}
\usepackage{xcolor}
\usepackage{cancel}
\usepackage{gensymb}
\usepackage[normalem]{ulem}
\usepackage{authblk}

\usepackage{mysymbols}

\usepackage{enumitem}
\usepackage[olditem,oldenum]{paralist}
\setdefaultleftmargin{.2cm}{.2cm}{}{}{}{}
\setlist{leftmargin=.2cm}

\usepackage[breaklinks=true,bookmarks=false]{hyperref}

\cvprfinalcopy 

\def\cvprPaperID{3450} 

\begin{document}

\title{Towards VQA Models That Can Read}
\author{Amanpreet Singh\textsuperscript{1}, Vivek Natarajan, Meet Shah\textsuperscript{1}, Yu Jiang\textsuperscript{1}, Xinlei Chen\textsuperscript{1},\\ Dhruv Batra\textsuperscript{1,2}, Devi Parikh\textsuperscript{1,2}, Marcus Rohrbach\textsuperscript{1}}
\affil{\textsuperscript{1}Facebook AI Research, \textsuperscript{2}Georgia Institute of Technology \\ \vspace{0.7em} \href{https://textvqa.org}{https://textvqa.org}}

\newcommand{\datasetName}{TextVQA\xspace}
\newcommand{\datasetNQuestions}{45,336\xspace}
\newcommand{\datasetNImages}{28,408\xspace}
\newcommand{\sectionReduceTop}{-5mm}

\newcommand{\approachName}{\emph{Look, Read, Reason \& Answer} (LoRRA)\xspace}
\newcommand{\approachNameShort}{LoRRA\xspace}
\newcommand{\icon}[1]{\raisebox{-0.32\height}{\includegraphics[height=22pt]{#1}}}

\maketitle


\begin{abstract}
   Studies have shown that a dominant class of questions asked by visually impaired users on images of their surroundings involves reading text in the image. But today's VQA models can not read! Our paper takes a first step towards addressing this problem. First, we introduce a new \emph{``\datasetName''} dataset to facilitate progress on this important problem. Existing datasets either have a small proportion of questions about text (e.g., the VQA dataset) or are too small (e.g., the VizWiz dataset). \datasetName contains \datasetNQuestions{} questions on \datasetNImages{} images that require reasoning about text to answer.
   Second, we introduce a novel model architecture that reads text in the image, reasons about it in the context of the image and the question, and predicts an answer which might be a deduction based on the text and the image or is composed of the strings found in the image. Consequently, we call our approach \approachName\footnote{Code is available at \href{https://github.com/facebookresearch/pythia}{https://github.com/facebookresearch/pythia}}.
   We show that \approachNameShort~ outperforms existing state-of-the-art VQA models on our \datasetName dataset. 
   We find that the gap between human performance and machine performance is significantly larger on TextVQA than on VQA 2.0, suggesting that TextVQA is well-suited to benchmark progress along directions complementary to VQA 2.0. 
\end{abstract}

\section{Introduction}
\label{sec:intro}
The focus of this paper is endowing 
Visual Question Answering (VQA) 
models a new capability -- the ability to \emph{read text in images and answer questions} by reasoning over the text and other visual content. 

\begin{figure}[t]
    \centering
    \includegraphics[width=0.46\textwidth, height=0.54\textwidth]{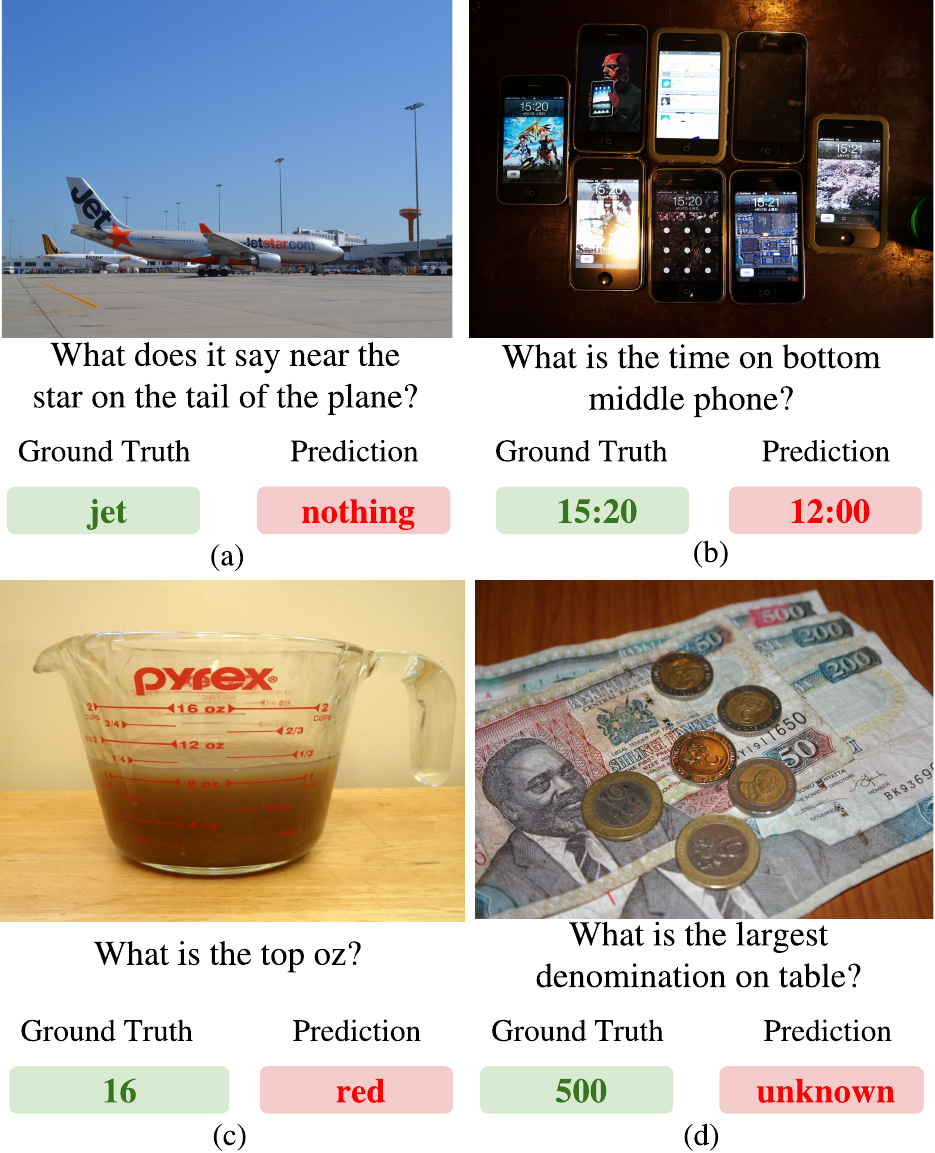}    
    \caption{
    \textbf{Examples from our \datasetName dataset}. \datasetName questions require VQA models to understand text embedded in the images to answer them correctly. Ground truth answers are shown in green and the answers predicted by a state-of-the-art VQA model (Pythia~\cite{jiang2018pythia}) are shown in red.
    Clearly, today's VQA models fail at answering questions that involve reading and reasoning about text in images.
    }
    \label{fig:teaser}
    \vspace{-5mm}
\end{figure}
VQA has witnessed tremendous progress. But today's VQA models fail catastrophically on questions 
requiring reading!\footnote{All top entries in the CVPR VQA Challenges (2016-18) struggle to answer questions in category requiring reading correctly.}
This is ironic because these are \emph{exactly} the questions 
visually-impaired users frequently ask of their assistive devices. Specifically, the VizWiz study~\cite{bigham2010vizwiz} found that up to 21\% of these questions
involve reading and reasoning about the text captured in the images of a user's surroundings -- 
\myquote{what temperature is my oven set to?}, 
\myquote{what denomination is this bill?}.

Consider the question in Fig.~\ref{fig:teaser}(a) -- 
\myquote{What does it say near the star on the tail of the plane?} from the \datasetName dataset. 
With a few notable exceptions, today's state-of-art VQA models are predominantly 
monolithic deep neural networks (without any specialized components). 
Consider what we are asking such models to learn; for answering these questions, the model must learn to 
\begin{compactitem}
    \item realize when the question is about text (\myquote{What \dots say?}), 
    \item detect image regions containing text (\myquote{15:20}, \myquote{500}),
    \item convert pixel representations of these regions 
    (convolutional features) to symbols (\myquote{15:20}) or 
    textual representations (semantic word-embeddings), 
    \item jointly reason about detected text and visual content, \eg 
    resolving spatial or other visual reference relations 
    (\myquote{tail of the plane \ldots on the back}) to focus on the correct regions.
    \item finally, decide if the detected text needs to be  `copy-pasted' as the answer (\eg~\myquote{16} in Fig.~\ref{fig:teaser} (c))
    or if the detected text informs the model about an answer in the 
    answer space (\eg~answering \myquote{jet}, in Fig.~\ref{fig:teaser}(a)).
\end{compactitem}

When laid out like that, it is perhaps unsurprising why today's models have not been able to make progress on questions requiring reading and reasoning about text in the images -- simply put, despite all the strengths 
of deep learning, it seems hopelessly implausible that all 
of the above skills will simply \emph{emerge} in a 
monolithic network all from the distant supervision of VQA accuracy. 

Fortunately, we can do more than just hope. 
Optical Character Recognition (OCR) is a mature 
sub-field of computer vision. A key thesis of our work 
is the following --  we should bake in 
inductive biases and specialized 
components (\eg OCR) into models to endow them with the different 
skills (\eg reading, reasoning) required by the all-encompassing task of VQA. 

Specifically, we propose a new VQA model that includes OCR as a module. We call it \approachName. Our model architecture incorporates the regions (bounding boxes) in the image containing text as entities to attend over (in addition to object proposals). It also incorporates the actual text  recognized in these regions 
(\eg \myquote{15:20}) 
as information (in addition to visual features) that the model learns to reason over. Finally, our model includes a mechanism to decide if the answer produced should be `copied' over from the OCR output (in more of a generation or slot-filling flavor), or should be deduced from the text (as in a standard discriminative prediction paradigm popular among existing VQA models). Our model learns this mechanism end-to-end. While currently limited in scope to OCR, our model is as an initial step towards endowing VQA models with the ability to reason over unstructured sources of external knowledge (in this case text found in a test image) and accommodate multiple streams of information flow (in this case predicting an answer from a pre-determined vocabulary or generating an answer via copy). 

One reason why there has been limited progress on VQA models that can read and reason about text in images is because such questions, while being a dominant category in real applications for aiding visually impaired users~\cite{bigham2010vizwiz}, are infrequent 
in the standard VQA datasets~\cite{antol2015vqa, balanced_vqa_v2, zhu2016visual7w} because they were not collected in the settings that mimic those of visually impaired users.
While the VizWiz dataset~\cite{gurari2018vizwiz} does contain data collected from visually impaired users,
the effective size of the dataset is small due to 58\% of the questions being ``unanswerable''. 
This makes it challenging to study the problem systematically, train effective models, or even draw sufficient attention to this important skill that current VQA models lack. 

To this end, we introduce the \datasetName dataset. It contains \datasetNQuestions{} questions asked by (sighted) humans on \datasetNImages{} images from the Open Images dataset~\cite{krasin2016openimages} from categories that tend to contain text \eg ``billboard", ``traffic sign", ``whiteboard". 
Questions in the dataset require reading and reasoning about text in the image. Each question-image pair has 
10 ground truth answers provided by humans. 

Models that do well on this dataset will not only need to parse the image and the question as in traditional VQA, but also read the text in the image, identify which of the text might be relevant to the question, and recognize whether a subset of the detected text can directly be the answer (e.g., in the case of \myquote{what temperature is my oven set to?}) or additional reasoning is required on the detected text to answer the question (e.g., \myquote{which team is winning?}).

Overall, our contributions are:

\begin{compactitem}
\item We introduce a novel dataset (\datasetName) containing questions which require the model to read and reason about the text in the image to be answered. 
\item We propose \approachName: a novel model architecture which explicitly reasons over the outputs from an OCR system when answering questions. 
\item \approachNameShort outperforms existing state-of-the-art VQA models on our \datasetName as well as VQA 2.0 dataset.
\end{compactitem}

\vspace{-2mm}
\section{Related work}
\vspace{-2mm}
\begin{figure*}
    \centering
    \includegraphics[width=.90\textwidth]{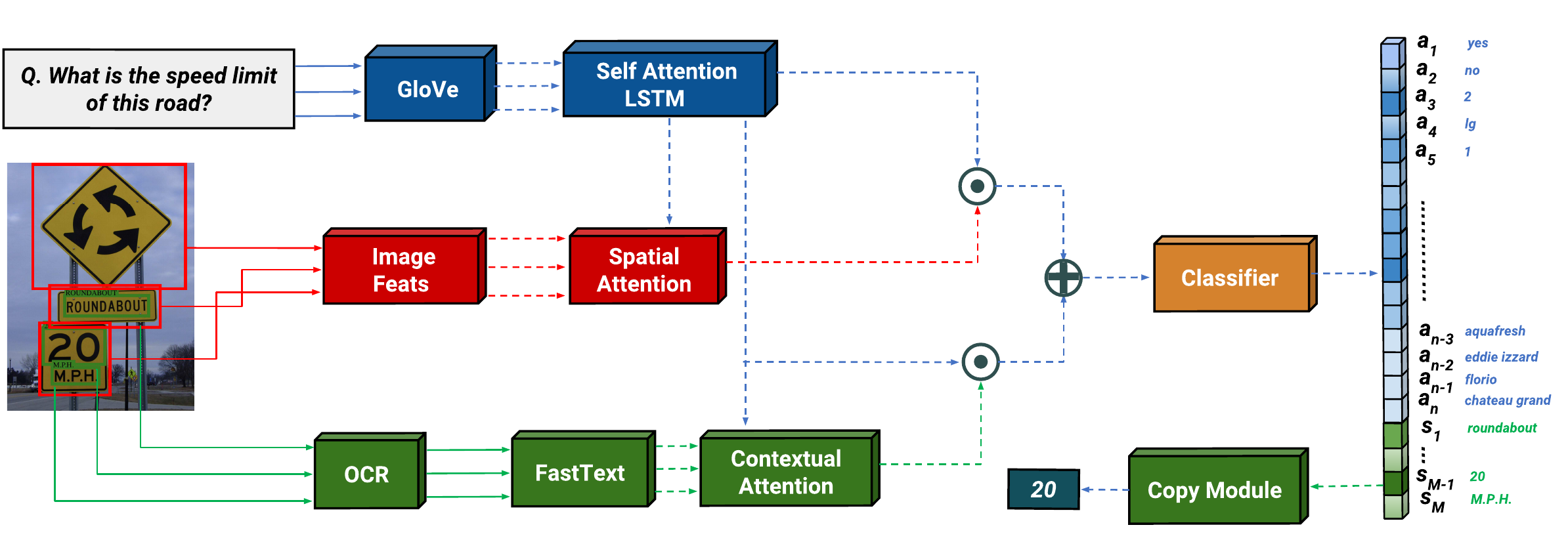}
    \caption{\textbf{Overview of our approach \approachName}. Our approach looks at the image, reads its text, reasons about the image and text content and then answers, either with an answer $a$ from the fixed answer vocabulary or by selecting one of the OCR strings $s$. Dashed lines indicate components that are not jointly-trained. The answer cubes on the right with darker color have more attention weight. The OCR token ``20'' has the highest attention weight in the example.
    }
    \label{fig:model}
    \vspace{-3mm}
\end{figure*}
\textbf{Visual Question Answering.} VQA has seen numerous advances and new datasets since the first large-scale VQA dataset was introduced by Antol~\etal~\cite{antol2015vqa}. 
This dataset was larger, more natural, and more varied than earlier VQA datasets such as DAQUAR \cite{malinowski2014multi} or COCO-QA \cite{ren2015exploring} but had linguistic priors which were exploited by models to answer questions without sufficient visual grounding. This issue was addressed by Goyal~\etal~\cite{balanced_vqa_v2} by adding complementary triplets ($I_c, q, a_c$) for each original triplet ($I_o, q, a_o$) where image $I_c$ is similar to image $I_o$ but the answer for the given question $q$ changes from $a_o$ to $a_c$. To study visual reasoning independent of language, non-photo-realistic VQA datasets have been introduced such as CLEVR \cite{johnson2017clevr}, NLVR \cite{suhr2017corpus} and FigureQA \cite{kahou2017figureqa}.
Wang~\etal~\cite{wang2018fvqa} introduced a Fact-Based VQA dataset which explicitly requires external knowledge to answer a question. 

\textbf{Text based VQA.} Several existing datasets study text detection and/or parsing in natural everyday scenes: COCO-Text \cite{veit2016coco}, Street-View text \cite{wang2010word} IIIT-5k \cite{mishra2012scene} and ICDAR 2015 \cite{karatzas2015icdar}. These do not involve answering questions about the images or reasoning about the text.
DVQA \cite{kafle2018dvqa} assesses automatic bar-chart understanding by training models to answer questions about graphs and plots. The Multi-Output Model (MOM) introduced in DVQA  uses an OCR module to read chart specific content. 
Textbook QA (TQA) \cite{kembhavi2017you} considers the task of answering questions from middle-school textbooks, which often require understanding and reasoning about text and diagrams. 
Similarly, AI2D \cite{kembhavi2016diagram} contains diagram based multiple-choice questions. MemexQA \cite{jiang2017memexqa} introduces a VQA task which involves reasoning about the time and date at which a photo/video was taken, but this information is structured and is part of the meta data.
Note that these works all require reasoning about text to answer questions, but in narrow domains (bar charts, textbook diagrams, etc.). The focus of our work is to reason and answer questions about text in natural everyday scenes. 

\textbf{Visual Representations for VQA Models.}
VQA models typically use some variant of attention to get a representation of the image that is relevant for answering the given question \cite{andreas2016neural, fukui2016multimodal,lu2016hierarchical,xu2016ask,yang2016stacked,zhu2016visual7w, jiang2018pythia}.
The object region proposals and the associated features are generated by using a detection network which are then spatially attended to and conditioned on a question representation.
In this work, we extend the representations that a VQA model reasons over. Specifically, in addition to attending over
object proposals, our model also attends over the regions where text is detected. 

\textbf{Copy Mechanism.}
A core component of our proposed model is its ability to decide whether the answer to a question should be an OCR token detected in the image, or if the OCR tokens should only inform about the answer to the question. 
The former is implemented as a ``copy mechanism'' -- a learned slot filling approach.
Our copy mechanism is based on a series of works on
the pointer generator networks \cite{gu2016incorporating,see2017get,merity2016pointer,gulcehre2016pointing,nallapati2016abstractive}. 
A copy mechanism provides networks the ability to generate out-of-vocabulary words by pointing at a word in context and then copying it as the answer. This approach has been used for a variety of tasks in NLP such as summarization \cite{gu2016incorporating, nallapati2016abstractive, see2017get}, question answering \cite{xiong2016dynamic}, language modelling \cite{merity2016pointer}, neural machine translation \cite{gulcehre2016pointing}, and dialog \cite{raghu2018hierarchical}.
\vspace{-2mm}
\section{\approachNameShort:
Look, Read, Reason \& Answer}
\label{sec:model}
\vspace{-1mm}
In this section, we introduce our novel model architecture to answer  questions which require reading text in the image.

We assume we get an image $v$ and a question $q$ as the input, where the question consists of $L$ words $w_1,w_2,\ldots,w_L$. At a high level, our model contains three components: (i) a \textbf{VQA component} to reason and infer about the answer based on the image $v$ and the question $q$ (Sec~\ref{subsec:answer_module}); (ii) a \textbf{reading component} which allows our model to read the text in the image (Sec~\ref{subsec:reading}); and (iii) an \textbf{answering module} which either predicts from an answer space or points to the text read by the \textit{reading component} (Sec.~\ref{subsec:answer_module}). The overall model is shown in Fig.~\ref{fig:model}. Note that, the OCR module and backbone VQA model can be any OCR model and any recent attention-based VQA model. Our approach is agnostic to the internal details of these components. 
We detail our exact implementation choices and hyper parameters in Sec.~\ref{subsec:implementation}. 

\subsection{VQA Component}
\label{subsection:vqa_component}
Similar to many VQA models \cite{fukui2016multimodal,jiang2018pythia}, we first embed the question words $w_1,w_2,\ldots,w_L$ of the question $q$ with a pre-trained embedding function (\eg GloVe \cite{pennington2014glove}) and then encode the resultant word embeddings iteratively with a recurrent network (\eg LSTM \cite{hochreiter1997long}) to produce a question embedding $f_Q(q)$. For images, the visual features are represented as spatial features, either in the form of grid-based convolutions and/or features extracted from the bounding box proposals~\cite{anderson2017bottom}. We refer to these features as $f_I(v)$ where $f_I$ is the network which extracts the image representation.
We use an attention mechanism $f_A$ over the spatial features~\cite{bahdanau2014neural,fukui2016multimodal}, which predicts attentions based on the $f_I(v)$ and $f_Q(q)$ and gives a weighted average over the spatial features as the output. 

We then combine the output with the question embedding. At a high level, the calculation of our VQA features $f_{VQA}(v, q)$ can be written as:
\begin{equation}
    f_{VQA}(v,q)= f_{comb}(f_A(f_I(v), f_Q(q)), f_Q(q))
\end{equation}
where $f_{comb}$ is the combination module ($\bigotimes$) in Fig.~\ref{fig:model}.

Assuming that we have a fixed answer space of $a_1,\ldots, a_N$, we use a feed-forward MLP $f_c$ on the combined embedding $f_{VQA}(v, q)$ to predict probabilities $p_1, \ldots, p_N$ where the probability of $a_i$ being the correct answer is $p_i$. 

\subsection{Reading Component}
\label{subsec:reading}
To add the capability of reading text from an image, we rely on an OCR model which is not jointly trained with our system. We assume that the OCR model can read and return word tokens from an image, e.g. \cite{borisyuk2018rosetta,smith2007overview}. The OCR model extracts $M$ words $s=s_1,s_2, ..., s_M$ from the image which are then embedded with a pre-trained word embedding, $f_O$. Finally, we use the same architecture as VQA component to get combined OCR-question features, $f_{OCR}$. Specifically, 
\begin{equation}
    f_{OCR}(s, q)= f_{comb}(f_A(f_O(s), f_Q(q)), f_Q(q))
\end{equation}
This is visualized in Fig.~\ref{fig:model}.
Note that the parameters of the functions $f_A$ and $f_{comb}$ are not shared with the VQA model component above but they have the same architecture, just with different input dimensions. 

During weighted attention because the features are multiplied by weights and then averaged, the ordering information gets lost. To provide the answer module with the ordering information of the original OCR tokens, we concatenate the
attention weights with the final weight-averaged features. 
This allows the answer module to know the original attention weights for each token in order. 
\begin{figure*}[ht]
    \centering
    \footnotesize
    \includegraphics[width=1\textwidth]{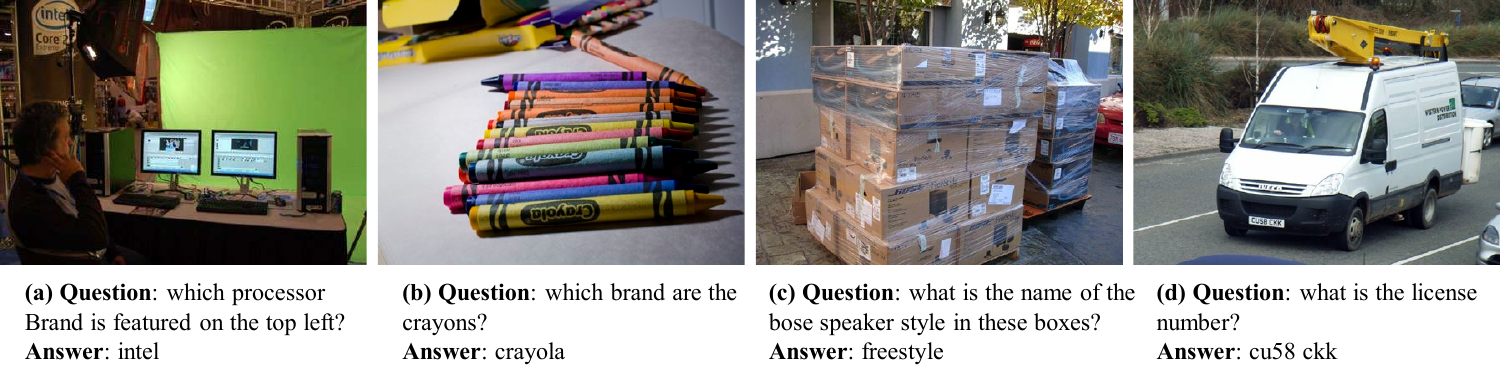}
    \caption{\textbf{Examples from \datasetName.} Questions require inferring hidden characters (``intel"), handling rotated text (``crayola"), reasoning (``bose" versus ``freestyle") and selecting among multiple texts in image ``cu58 ckk" versus ``western power distribution").
    }
    \label{fig:annotation_examples}
    \vspace{-5mm}
\end{figure*}

\subsection{Answer Module}
\label{subsec:answer_module}

With a fixed answer space, the current VQA models are only able to predict fixed tokens which limits the generalization to out-of-vocabulary (OOV) words. As the text in images frequently contains words not seen at 
training time, it is hard to answer text-based questions based on a pre-defined answer space alone.
To generalize to arbitrary text,
we take inspiration from pointer networks which allow pointing to OOV words in context \cite{gu2016incorporating,see2017get,merity2016pointer,gulcehre2016pointing,nallapati2016abstractive}. We extend our answer space through addition of a dynamic component which corresponds to $M$ OCR tokens. The model now has to predict probabilities $(p_1, \ldots, p_N, \ldots, p_{N+M})$ for $N+M$ items in the answer space instead of the original $N$ items. 

We pick the index with the highest probability $p_i$ as the index of our predicted answer. If the model predicts an index larger than $N$ (i.e., among the last $M$ tokens in answer space), we directly \textit{``copy"} the corresponding OCR token as the predicted answer. Hence, our answering module can be thought of as \textit{``copy if you need"} module which allows answering from the OOV words using the OCR tokens.

With all of the components, the final equation $f_{\approachNameShort}$ for predicting the answer probabilities
can be written as: 
\begin{equation}
    f_{\approachNameShort}(v,s,q)=f_{MLP}([f_{VQA}(v, q); f_{OCR}(s, q)])
\end{equation}
where $[;]$ refers to concatenation and $f_{MLP}$ is a two-layer feed-forward network which predicts the binary probabilities as logits for each answer. We opt for binary cross entropy using logits instead of calculating the probabilities through softmax as it allows us to handle cases where the answer can be in both the actual answer space and the OCR tokens without penalizing for predicting either one (the likelihood of logits is independent of each other). Note that if the model chooses to copy, it can only produce one of the OCR tokens as the predicted answer. 8.9\% of the \datasetName questions can only be answered by combining multiple OCR tokens; we leave this as future work.

\begin{table}[t]
\centering
\setlength\tabcolsep{8pt}
\footnotesize
\renewcommand{\arraystretch}{1.2}
\begin{tabular}{@{}p{4.9cm}r@{}}
\toprule
\multicolumn{2}{c}{\textbf{VQA 2.0 Accuracy}} \\
\textbf{Model} & \textbf{test-dev}
\\
\midrule
\textbf{BUTD~\cite{anderson2017bottom}}                                 & 65.32  
\\
\textbf{Counter~\cite{zhang2018counting}}                        & 68.09 
\\  
\textbf{BAN \cite{kim2018bilinear}}               & 69.08 

\\
\textbf{Pythia v0.1 \cite{jiang2018pythia}}                                  & 68.49 
\\
\textbf{Pythia v0.3 (Ours)}                                                           & 68.71
\\
\textbf{Pythia v0.3 + LoRRA (Ours)}                                                           & 69.21
\\
\midrule
\multicolumn{2}{c}{\textbf{VizWiz Accuracy}} \\
\textbf{Model} & \textbf{test}  \\
\midrule
\textbf{BAN[25]} & 51.40 \\
\textbf{Pythia v0.3 (Ours)} & 54.72 \\
\bottomrule

\end{tabular}
\caption{\textbf{Single model VQA 2.0 and VizWiz performance in \%}. 
Our revised implementation of Pythia, v0.3, with LoRRA outperforms or is comparable to state-of-the-art on VQA 2.0.}
\vspace{-6mm}
\label{tab:vqaaccuracy}
\end{table}
\subsection{Implementation Details}
\label{subsec:implementation}
Our VQA component is based on the VQA 2018 challenge winner entry, Pythia v0.1 \cite{jiang2018pythia}. Our revised 
implementation, Pythia v0.3 \cite{singhpythia}, with slight changes in hyper-parameters (\eg size of question vocabulary, hidden dimensions) achieves state-of-the-art VQA accuracy for a single model (\ie w/o ensemble) as shown in Tab.~\ref{tab:vqaaccuracy} on both VQA v2.0 dataset~\cite{goyal2017accurate} and VizWiz dataset~\cite{gurari2018vizwiz}. The revised design choices are discussed in~\cite{singhpythia}.

Pythia \cite{jiang2018pythia,singhpythia} is 
inspired from the detector-based bounding box prediction approach of the bottom-up top-down attention network \cite{anderson2017bottom} (VQA winner 2017), which in turn has a multi-modal attention mechanism similar to the VQA 2016 winner \cite{fukui2016multimodal}, which relied on  grid-based features.

In Pythia, for spatial features $f_I(v)$, we rely on both grid and region based features for an image. The grid based features are obtained by average pooling 2048$\mathcal{D}$ features from the \texttt{res-5c} block of a pre-trained ResNet-152~\cite{he2016deep}. The region based features are extracted from the \texttt{fc6} layer of an improved Faster-RCNN model~\cite{Detectron2018} trained on the Visual Genome~\cite{krishna2017visual} objects and attributes as provided in \cite{anderson2017bottom}. During training, we fine-tune the \texttt{fc7} weights as in \cite{jiang2018pythia}.

We use pre-trained GloVe embeddings with a custom vocabulary (top $\sim$77k question words in the VQA 2.0) for the question embedding \cite{pennington2014glove}. The $f_Q$ module passes GloVe embeddings to an LSTM \cite{hochreiter1997long} with self-attention \cite{yu2018beyond} to generate question's sentence embedding. For OCR, we run the Rosetta OCR system \cite{borisyuk2018rosetta} to provide us word strings $s_1,...,s_N$.
OCR tokens are first embedded using pretrained FastText embeddings ($f_O$) \cite{joulin2016bag}, which can generate word embeddings even for OOV tokens as explained in \cite{joulin2016bag}.

In $f_A$, the question embedding $f_Q(q)$ is used to obtain the top-down \ie task-specific attention on both $f_O(s)$ OCR tokens features and $f_I(v)$ image features. The features are then averaged based on the attention weights to get a final feature representation for both the OCR tokens and the image features. The final grid-level and region-based features are concatenated in case of the image features. For the OCR tokens, attention weights are concatenated to the final attended features as explained in Sec.~\ref{subsection:vqa_component}. Finally, in $f_{comb}(x, y)$, the two feature embeddings in consideration are fused using element-wise/hadamard product, $\otimes$, of the features. 
The fused features from $f_{OCR}(s, q)$ and $f_{VQA}(v, q)$ are concatenated and passed through an MLP to produce logits from which word corresponding to maximum logit's index is selected as the answer. 

\vspace{-3mm}
\section{\protect\icon{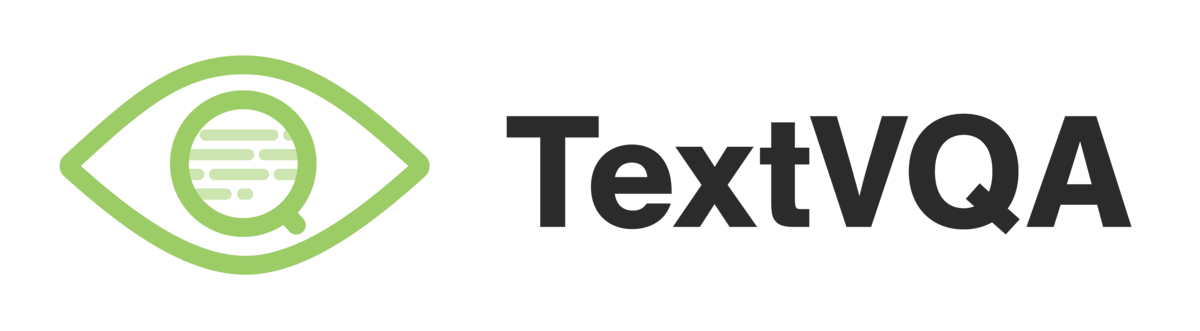}}
\begin{figure}
    \centering
    \includegraphics[width=1\linewidth, angle=0]{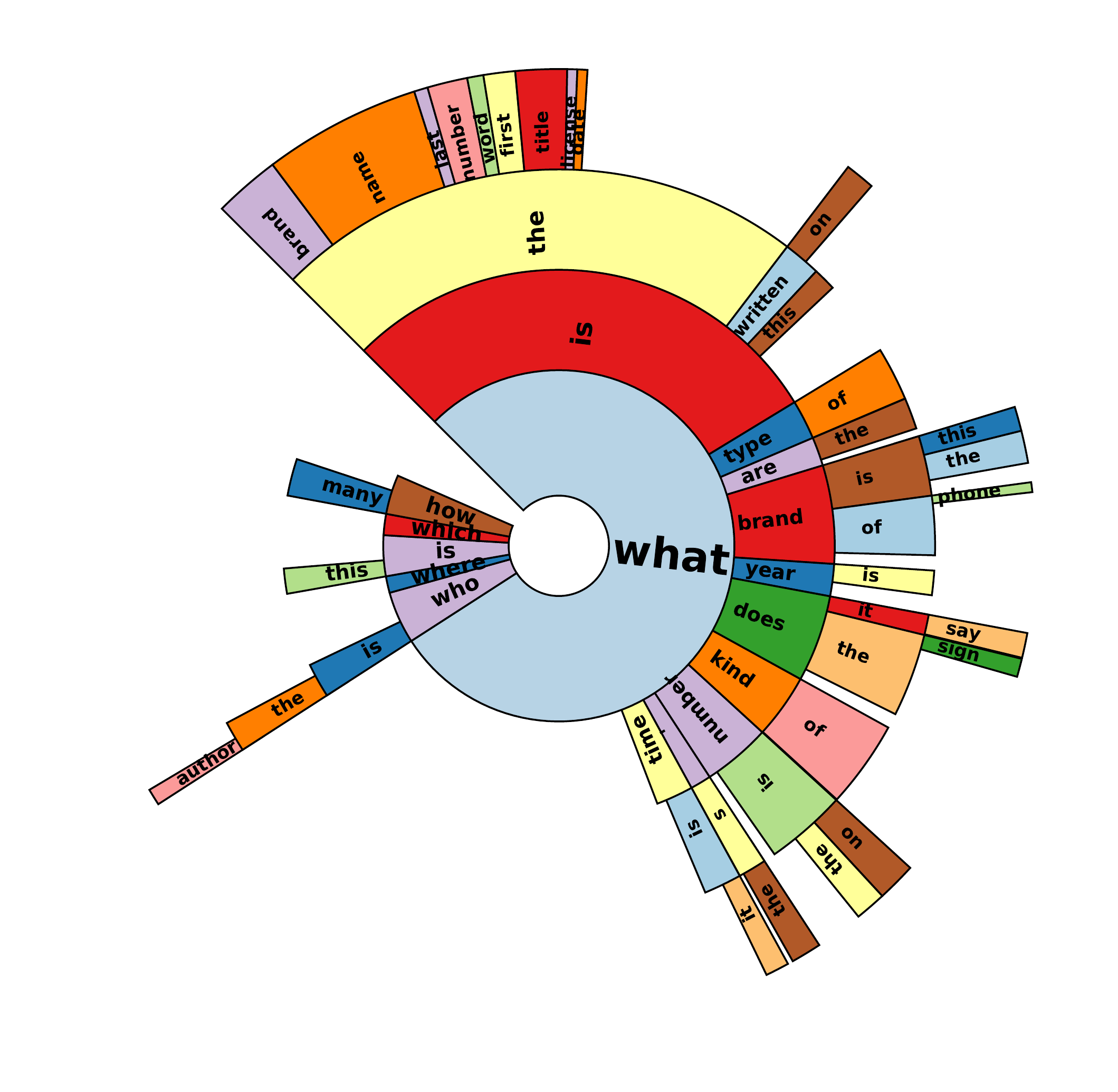}
    \vspace{-30pt}
    \caption{\textbf{Distribution of first four words in questions in \datasetName.} Most questions start with ``what''. 
    }
    \label{fig:sunburst}
    \vspace{-5pt}
\end{figure}

\begin{figure*}
    \centering
    \begin{subfigure}[t]{0.32\textwidth}
        \includegraphics[width=1\linewidth]{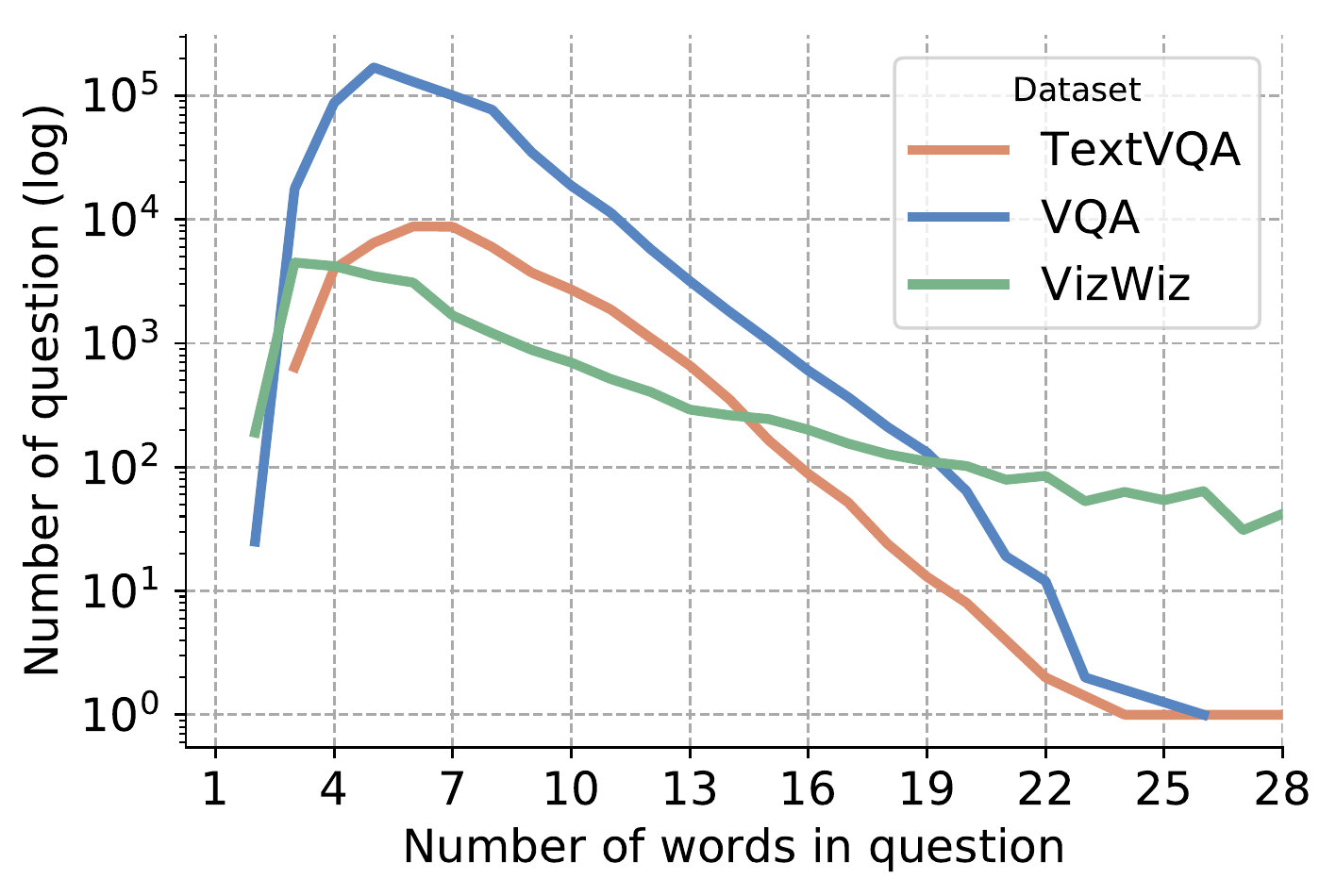}
        \caption{\textbf{Number of questions with a particular question length}. 
        We see that the average question length (7.16) is higher in \datasetName compared to others.}
        \label{fig:question_lengths}
    \end{subfigure}
    \hspace{1mm}
    \begin{subfigure}[t]{0.32\textwidth}
        \includegraphics[width=1\linewidth]{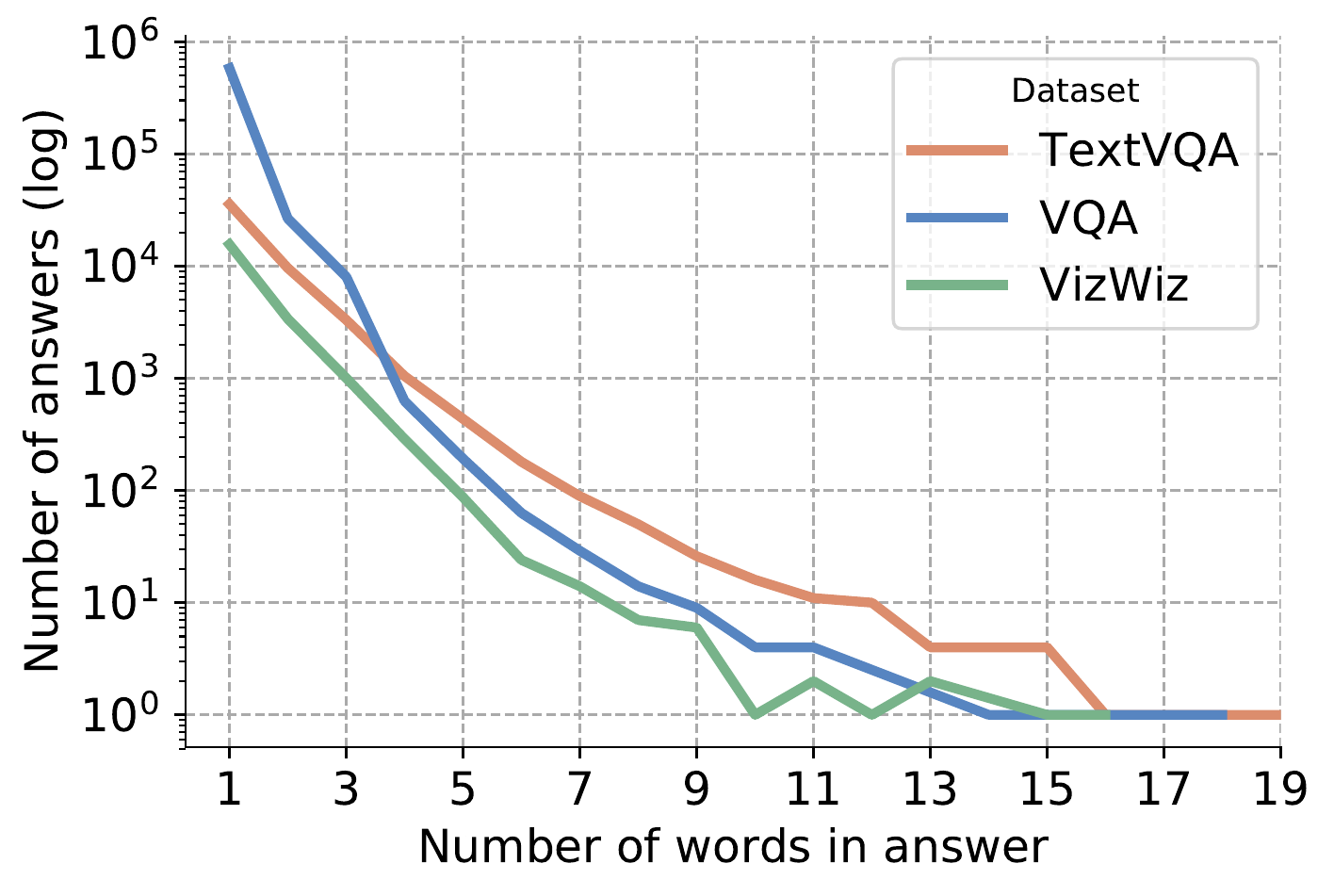}
        \caption{
        \textbf{Number of majority answers with a particular length}. Average answer length (1.7) is high and answer can contain long paragraph and quotes.}
        \label{fig:answer_lengths}
    \end{subfigure}
    \hspace{1mm}
    \begin{subfigure}[t]{0.32\textwidth}
        \includegraphics[width=1\linewidth]{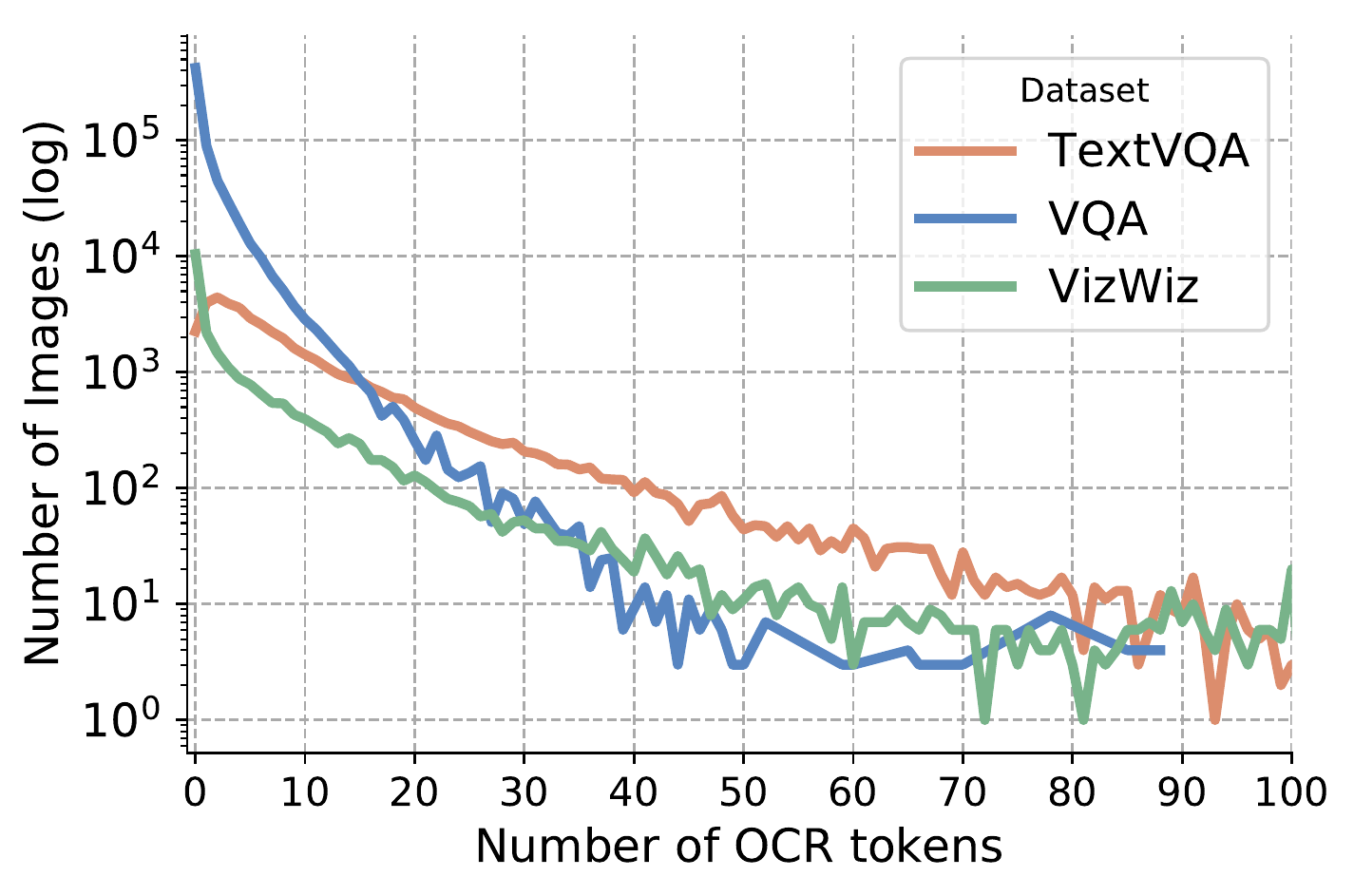}
        \caption{\textbf{Number of images with a particular number of OCR tokens}. Average number of tokens is around 3.14. In \datasetName, 10x more images contain OCR text than the others.}
        \label{fig:ocr_lengths}
    \end{subfigure}
    \begin{subfigure}[t]{0.32\textwidth}
        \includegraphics[width=1\linewidth]{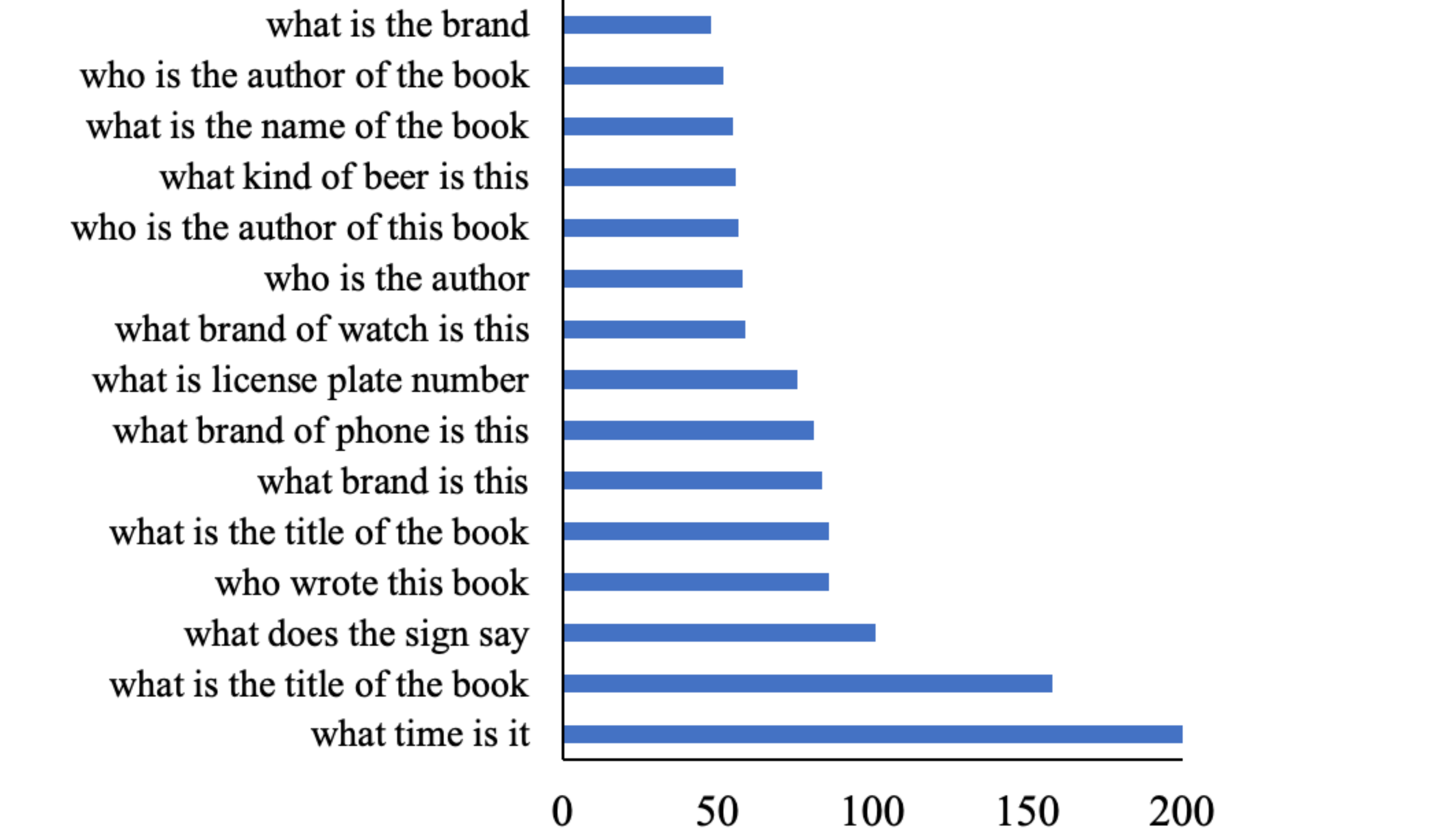}
        \caption{
        \textbf{Top 15 most occurring questions in \datasetName.} Most of the top questions start with ``what''.}
        \label{fig:question_count}
    \end{subfigure}
    \hspace{1mm}
    \begin{subfigure}[t]{0.32\textwidth}
        \includegraphics[width=1\linewidth]{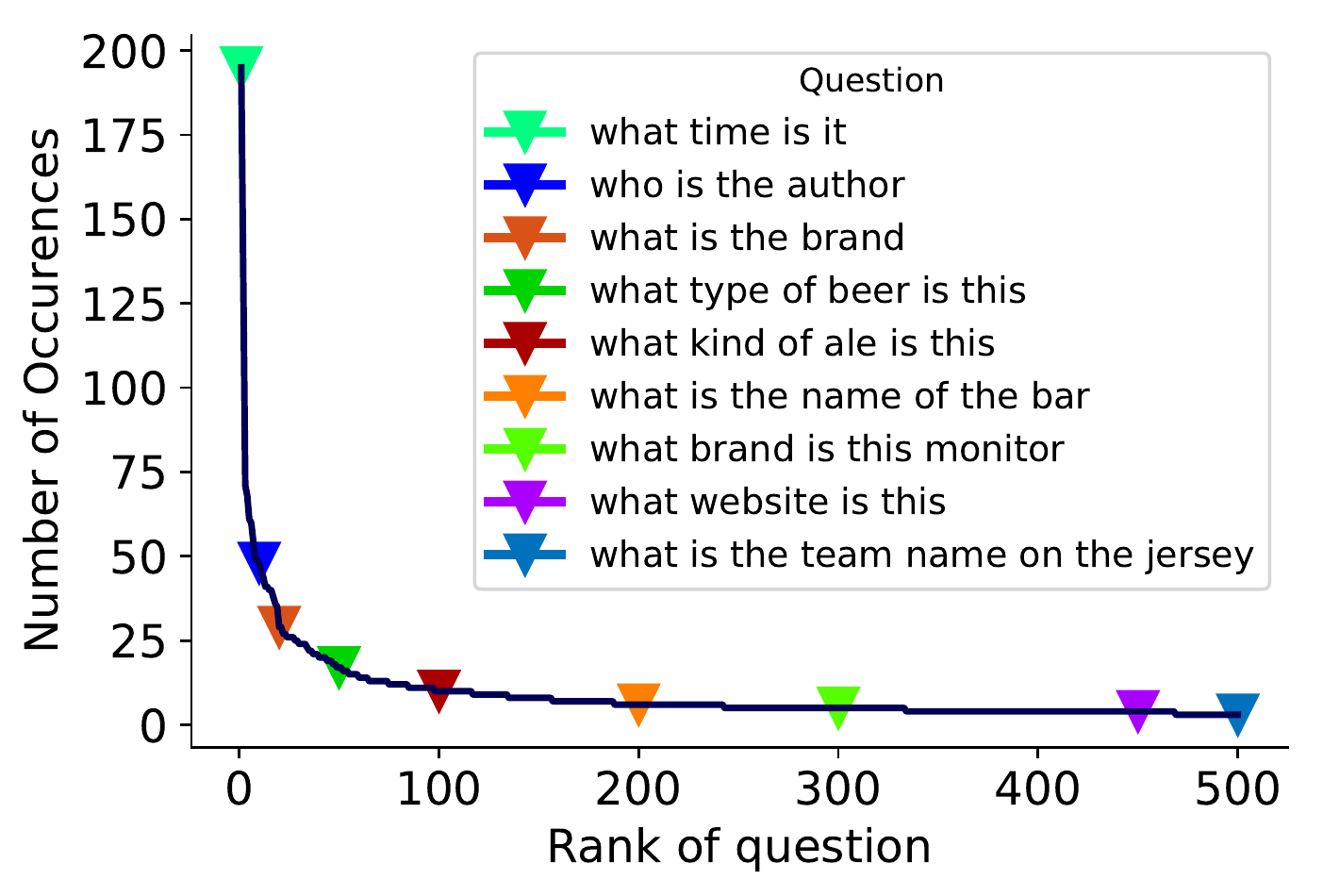}
        \caption{
        \textbf{Total occurrences for 500 most common questions} among 23184 unique questions with markers for particular ranks.}
        \label{fig:question_rank}
    \end{subfigure}
    \hspace{1mm}
    \begin{subfigure}[t]{0.32\textwidth}
        \includegraphics[width=1\linewidth]{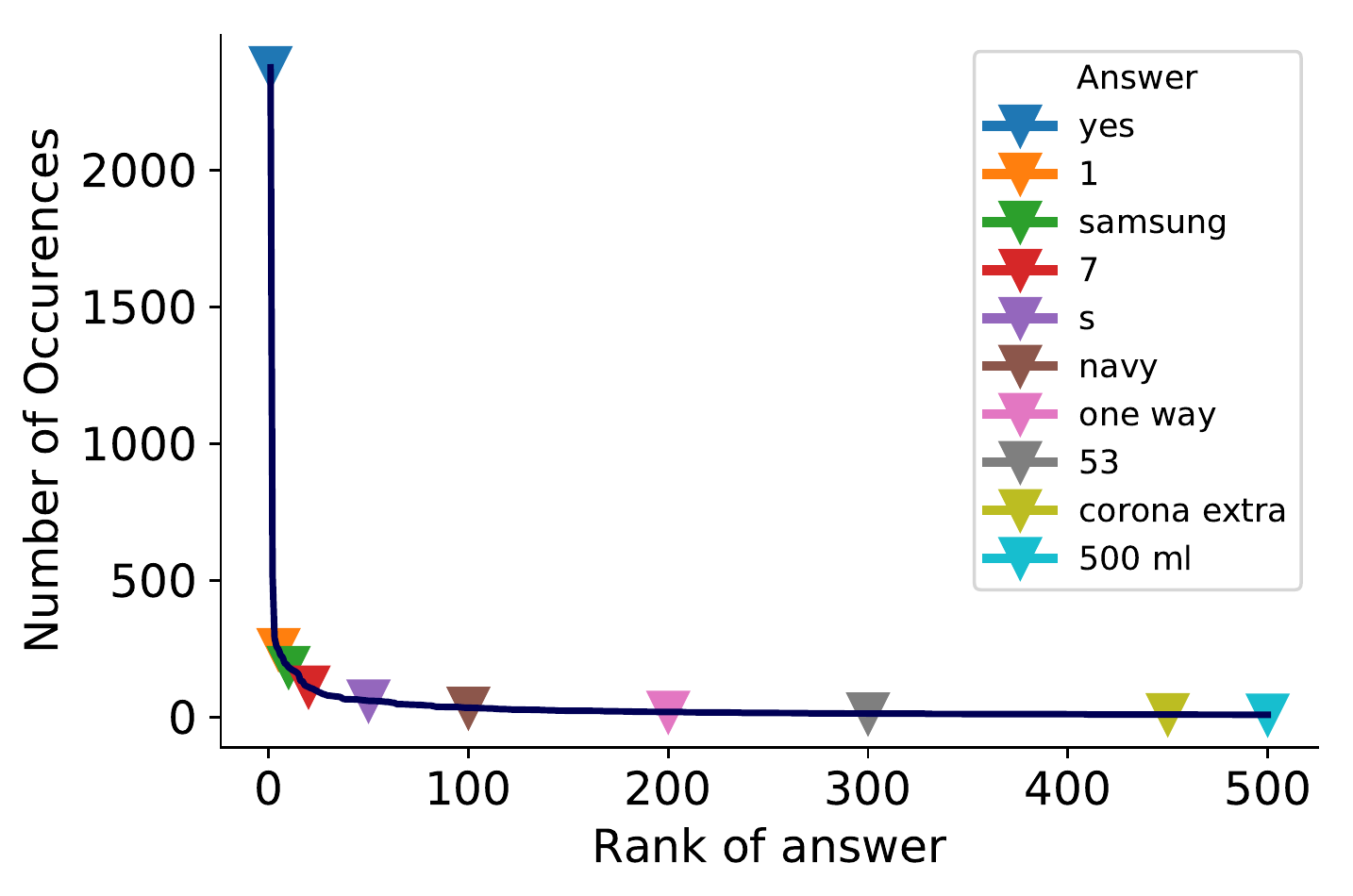}
        \caption{Similar to \ref{fig:question_rank}, plot shows \textbf{total occurrences for 500 most common majority answers} with markers for particular ranks.}
        \label{fig:answer_rank}
    \end{subfigure}
    \vspace{-3mm}
    \caption{\textbf{Question, Answer and OCR statistics for \datasetName}. We show comparisons with VQA 2.0 \cite{balanced_vqa_v2} and VizWiz \cite{gurari2018vizwiz}.}
    \vspace{-5mm}
\end{figure*}
To study the task of answering questions that require reading text in
images, we collect a new dataset called \datasetName~which is publicly available at \underline{\href{https://textvqa.org/}{https://textvqa.org}}. In this section, we start by describing how we selected 
the images that we use 
in \datasetName. We then explain our data collection pipeline for collecting the questions and the answers. Finally, we provide statistics and an analysis of the dataset. Snapshots of the annotation interface and detailed instructions can be found in the Appendix~\ref{appendx:analysis}.

\begin{figure}[ht]
    \centering
    \includegraphics[width=0.5\linewidth]{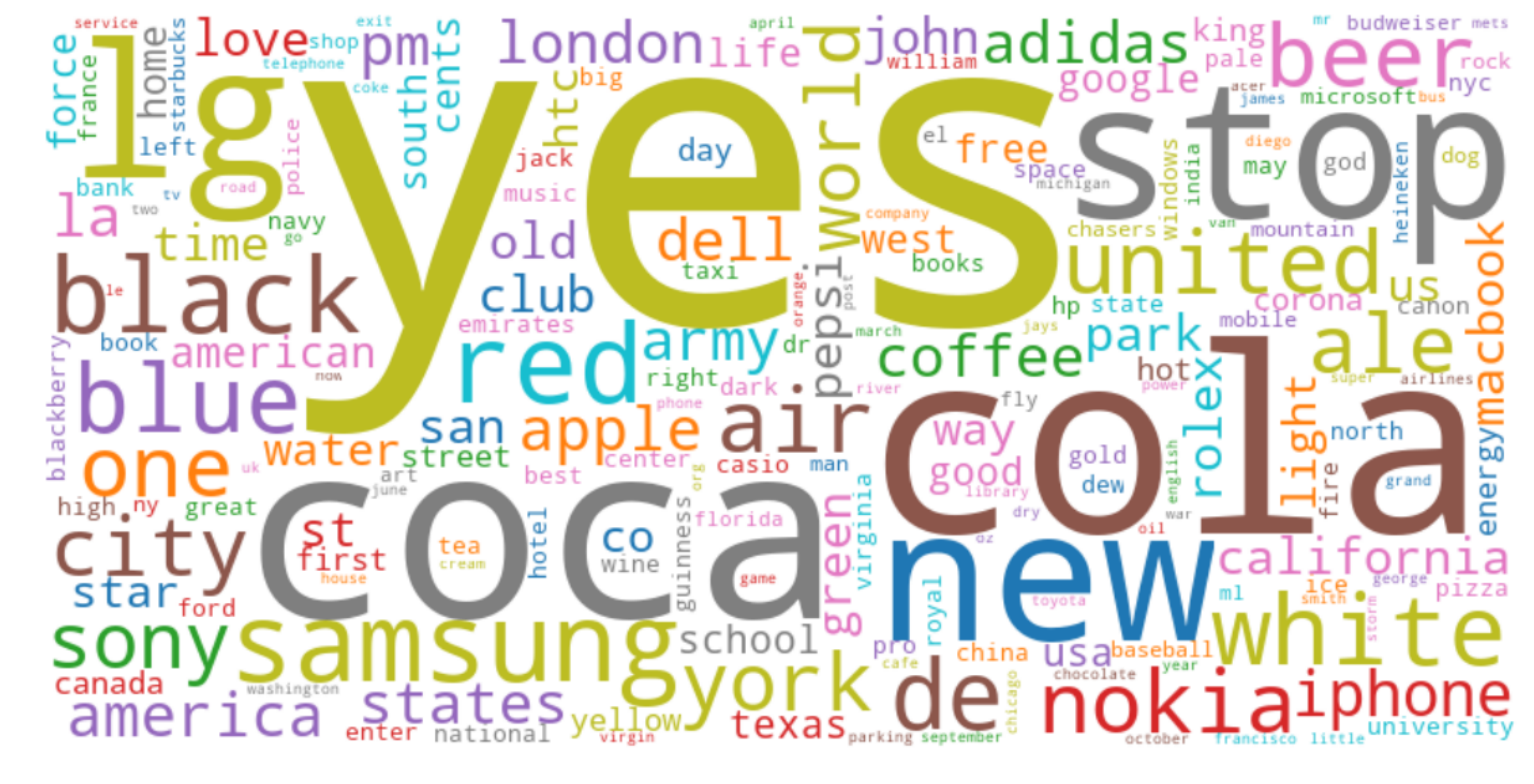}
    \includegraphics[width=0.49\linewidth]{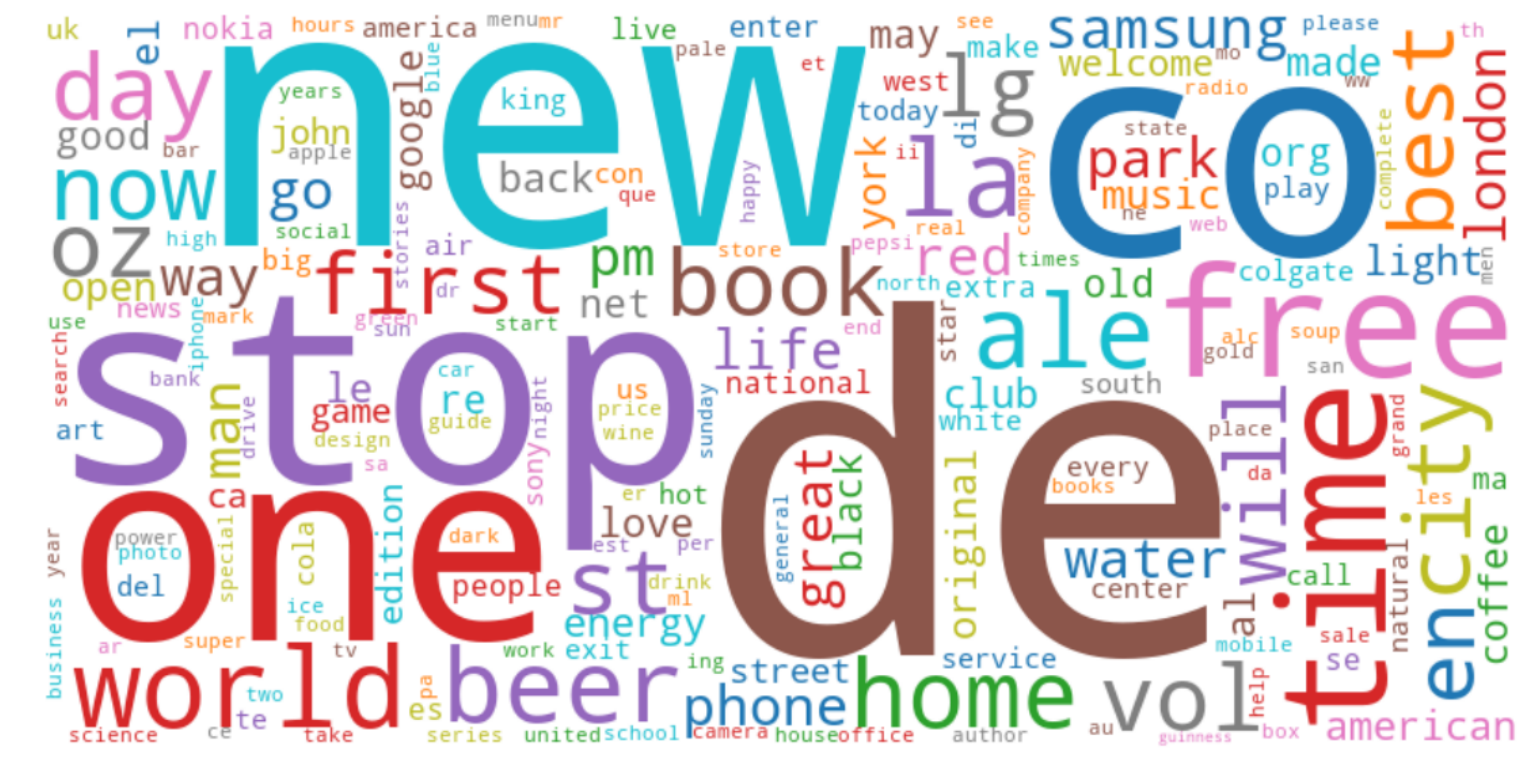}
    \caption{\textbf{(Left) Wordcloud for majority answers in \datasetName.} Frequently occurring answers include yes, brand names, ``stop'' and city names. \textbf{(Right) Wordcloud for OCR tokens predicted by Rosetta.} Note the overlap with answers on brand names \textit{(lg)}, cities \textit{(london)} and verbs \textit{(stop)}.
    }
    \label{fig:wordcloud}
    \vspace{-3mm}
\end{figure}

\subsection{Images}
We use Open Images v3 dataset~\cite{krasin2016openimages} as the source of our images. In line with the goal of developing and studying VQA models that can reason about text, we are most interested in the images that contain text in them. Several categories in Open Images fit this criterion (\eg, billboard, traffic  sign, whiteboard). To automate this process of identifying categories that tend to have images with text in them, we select 100 random images from each category (or all images if max images for that category is less than 100). We run a state-of-the-art OCR model Rosetta~\cite{borisyuk2018rosetta} on these images and compute the average number of OCR boxes in a category. The average number of OCR boxes per-category were normalized and used as per-category weights for sampling the images from the categories.

We collect \datasetName's training and validation set from Open Images' training set while test set is collected from Open Images' test set. We set up a three stage pipeline for crowd-sourcing our data. In the first stage, annotators were asked to identify images that did not contain text (using a forced-choice ``yes''/``no'' flag). Filtering those (and noisy data from annotators) out resulted in \datasetNImages images, which form the basis of our \datasetName{} dataset. 

\subsection{Questions and Answers}
In the second stage, we 
collect 1-2
questions for each image. For the first question, we show annotators an image and ask them to provide a question which requires reading the text in the image to answer. Specifically, they were told to \myquote{Please ensure that answering the question requires reading of the text in the image. It is OK if the answer cannot be directly copied from the text but needs to be inferred or paraphrased.} 

To collect a second question that is different from the first, we show annotators the first question and ask them to come up with a question 
that requires reasoning about the text in the image and has a different answer.
Following VQA~\cite{antol2015vqa,balanced_vqa_v2} and VizWiz~\cite{gurari2018vizwiz} datasets, we collect 10 answers for each question.

To ensure answer quality, we gave annotators instructions similar to those used in \cite{antol2015vqa,gurari2018vizwiz} when collecting the VQA and VizWiz datasets.
In addition, to catch any poor quality data from earlier steps,
we give annotators these four options: 
(i) no text in image; (ii) not a question; (iii) answering the question doesn't require reading any text in image; and (iv) unanswerable, \eg questions 
involving speculation about the meaning of text.
We remove the questions where a majority of  workers marked any of these flags. Additionally, we use hand-crafted questions for which we know the correct answers to identify and filter out bad annotators. 

\subsection{Statistics and Analysis}

We first analyze the diversity of the questions that we have in the dataset. \datasetName contains 45,336 questions of which 37,912 (83.6\%) are unique. 
Fig.~\ref{fig:question_lengths} shows the distribution of question length along with the same statistics for the VQA 2.0 and the VizWiz datasets for reference.
The average question length in \datasetName is 7.18 words which is higher than in  VQA 2.0 (6.29) and VizWiz (6.68). We also note that the minimum question length is 3 words. Workers often form questions which are longer to disambiguate the response (\eg specifying where exactly the text is in the image, see Fig.~\ref{fig:annotation_examples}).
Fig.~\ref{fig:question_count} shows top 15 most occurring questions in the dataset with their count while Fig.~\ref{fig:question_rank} shows top 500 most occurring questions 
with their counts. We can see the uniform shift from common questions about ``time'' to questions occurring in specific situations like ``team names''. Fig.~\ref{fig:sunburst} shows sunburst for first 4 words in questions. 
We also observe that most questions involve reasoning about common things (\eg figuring out brand names, cities and temperature). 
Questions often start with ``what'', 
frequently inquiring about ``time'', ``names'', ``brands'' or ``authors''.

In total there are 26,263 (49.2\%) unique majority answers in \datasetName. The percentage of unique answers in \datasetName is quite high compared to VQA 2.0 (3.4\%) and VizWiz (22.8\%). All 10 annotators agree on the most common answer for 22.8\% questions, while 3 or more annotators agree on most common answer for 97.9\% questions. Fig.~\ref{fig:wordcloud} (left) shows a word cloud plot for the majority answers in the dataset. The answer space is diverse and involves brand names, cities, people's names, time, and countries. Note that this diversity makes it difficult to have a fixed answer space -- a challenge that most existing VQA datasets do not typically pose. The most common answer (``yes") is the majority answer for only 4.71\%  of the dataset and ``yes/no" (majority answer) questions in total only make up 5.55\% of the dataset. The average answer length is 1.58 (Fig.~\ref{fig:answer_lengths}).
In a few occurrences where the text in the image is long (e.g., a quote or a paragraph), the answer length is high.  Fig. \ref{fig:answer_rank} shows the frequency of top 500 most common answers. The gradual shift from brands to rare cities is depicted. We also note that the drop in \datasetName for number of answers of a particular answer length is more gradual than in VQA 2.0 which drops sharply after answer length 3.

\begin{table}[t]
    \center
    \setlength{\tabcolsep}{0.5pt}
    \footnotesize
    
    \begin{tabular}{@{}lllr@{}r@{}}
    \toprule
    \multicolumn{3}{l}{} &  \multicolumn{2}{r}{\textbf{Accuracy(\%)}} \\
    \cmidrule{4-5}
    \multicolumn{3}{l}{\textbf{Model}} & \multicolumn{1}{r}{\textbf{Val}} & \multicolumn{1}{r}{\textbf{Test}}\\
    \addlinespace[0.5mm]
    \midrule
    \multicolumn{3}{l}{\textbf{Human}} & 85.01 & 86.79\\
    \multicolumn{3}{l}{\textbf{OCR UB}} & 37.12 & 36.52\\
    \multicolumn{3}{l}{\textbf{LA UB}} & 48.46 & 48.16\\
    \multicolumn{3}{l}{\textbf{LA+OCR UB}} & 67.56 & 68.24 \\
    \multicolumn{3}{l}{\textbf{Rand 100}} & 0.22 & 0.20\\
    \multicolumn{3}{l}{\textbf{Wt. Rand 100}} & 0.27 & 0.26\\
    \multicolumn{3}{l}{\textbf{Majority Ans}} & 4.48 & 2.63\\
    \multicolumn{3}{l}{\textbf{Random OCR}} & 7.72 & 9.12\\
    \multicolumn{3}{l}{\textbf{OCR Max}} & 9.76 & 11.60\\
    \addlinespace[0.5mm]
    \bottomrule
    \end{tabular}
     \:
    \begin{tabular}{@{}llllr@{}r@{}}
    \toprule
    \multicolumn{3}{l}{} & \multicolumn{1}{c}{} & \multicolumn{2}{c}{\textbf{Accuracy(\%)}} \\
    \cmidrule{5-6}
    \multicolumn{3}{l}{\textbf{Model}} & \multicolumn{1}{c}{\textbf{Vocab}} & \multicolumn{1}{r}{\textbf{Val}} & \multicolumn{1}{r}{\textbf{Test}} \\
    \midrule
    \multicolumn{3}{l}{\textbf{Q}} & \multicolumn{1}{c}{LA} & 8.09 & 8.70\\
    \multicolumn{3}{l}{\textbf{I}} & \multicolumn{1}{c}{LA} & 6.29 & 5.58\\
    
    \multicolumn{3}{l}{\textbf{Pythia (I+Q)}}& \multicolumn{1}{c}{LA} & 13.04 & 14.0\\
    \multicolumn{3}{l}{\textbf{~~~~~~+O }}& \multicolumn{1}{c}{LA} & 18.35 & -- \\
    \multicolumn{3}{l}{\textbf{~~~~~~+O+C }}& \multicolumn{1}{c}{n/a} & 20.06 & --\\
    \multicolumn{3}{l}{\textbf{~~~~~~+LoRRA }}& \multicolumn{1}{c}{LA} & 26.23 & --\\
    \multicolumn{3}{l}{\textbf{~~~~~~+LoRRA }}& \multicolumn{1}{c}{SA} & \textbf{26.56} & \textbf{27.63}\\ \hline
    \addlinespace[1mm]
    \multicolumn{3}{l}{\textbf{BAN (I+Q)}}& \multicolumn{1}{c}{LA} & 12.30 & --\\
    \multicolumn{3}{l}{\textbf{~~~~~~+LoRRA }}& \multicolumn{1}{c}{SA} & 18.41 & --\\
    \bottomrule
    \end{tabular}
    
    \caption{\textbf{Evaluation on \datasetName.} \textbf{(Left)} Accuracies for various heuristics baselines, which show that using \textbf{OCR} can help in achieving a good accuracy on \datasetName. \textbf{LA+OCR UB} refers to maximum accuracy achievable by models using \textbf{LoRRA} with our OCR tokens. 
    \textbf{(Right)} Accuracies of our trained baselines and ablations in comparison with our model \textbf{\approachNameShort}. \textbf{I} denotes usage of image features, \textbf{Q} question features,  \textbf{O}  OCR tokens' features, and \textbf{C} copy mechanism. \textbf{LA} and \textbf{SA} refer to use of large and short vocabulary, respectively. Models with \textbf{LoRRA} outperform VQA SoTA (Pythia, BAN) and other baselines.}
    \label{tab:results}
    \vspace{-18pt}
\end{table}

Finally, we analyze the OCR tokens produced by the Rosetta OCR system \cite{borisyuk2018rosetta}. In Fig. \ref{fig:ocr_lengths}, we plot number of images containing ``x'' number of OCR tokens. The peak between 4 and 5 shows that a lot of images in our dataset contain a good number of OCR tokens. In some cases, when the system is unable to detect text we get 0 tokens but those cases are restricted to $\sim$1.5k images and we manually verified that the images actually do contain text. 
Fig.~\ref{fig:wordcloud} (right) shows a word cloud of OCR tokens which shows they do contain common answers such as brand names and cities. 

\section{Experiments}
We start by explaining our baselines including both heuristics and end-to-end trained models which we compare with \approachNameShort.
We divide \datasetName into train, validation and test splits with size 34,602, 5,000,
and 5,734, respectively. 
The TextVQA questions collected from Open Images v3's training set were randomly split into training and validation sets. 
There is no image overlap between the sets.
For our approach, we use a vocabulary \textbf{SA} of size 3996, which contains answers which appear at least twice in the training set. For the baselines that don't use the copy mechanism, this vocabulary turns out to be too limited. To give them a fair shot, we also create a larger vocabulary \textbf{(LA)}, containing the 8000 most frequent answers.
\\

\noindent\textbf{Upper Bounds and Heuristics.}
These mainly evaluate the upper bounds of what can be achieved using the OCR tokens detected by our OCR module and benchmark biases in the dataset. We test (i) \textbf{OCR UB:} the upper bound accuracy one can get if the answer can be build directly from OCR tokens (and can always be predicted correctly). \textbf{OCR UB} considers combinations of OCR tokens upto 4-grams.
(ii) \textbf{LA UB:} the upper bound accuracy  by always predicting the  correct answer if it is present in \textbf{LA}. 
(iii) \textbf{LA+OCR UB:} (i) + (ii) - the upper bound accuracy one can get by predicting the correct answer if it is present in either LA or OCR tokens.
(iv) \textbf{Rand 100:} the accuracy one can get by selecting a random answer from top 100 most frequent answers (v) \textbf{Wt. Rand 100:} the accuracy of baseline (iv) but with weighted random sampling using 100 most occurring tokens' frequencies as weights. (vi) \textbf{Majority Ans:} the accuracy of always predicting the majority answer ``yes'' (vii) \textbf{Random OCR token:} the accuracy of predicting a random OCR token from the OCR tokens detected in an image (viii) \textbf{OCR Max:} accuracy of always predicting the OCR token that is detected maximum times in the image (e.g., ``crayola'' in Fig. \ref{fig:annotation_examples} (b)). 
\\

\noindent\textbf{Baselines.\footnote{Code is available at \underline{\href{https://github.com/facebookresearch/pythia}{https://github.com/facebookresearch/pythia}}}} We make modifications to the implementation discussed in Sec.~\ref{subsec:implementation} for our  baselines which include (i) \textbf{Question Only (Q):} we only use the $f_Q(q)$ module of \approachNameShort to predict the answer and the rest of the features are zeroed out. (ii) \textbf{Image Only (I):} similar to  \textbf{Q}, we only use image features $f_I(v)$ to predict answers.  
\textbf{Q} and \textbf{I} do not have access to OCR tokens and predict from \textbf{LA}. 
\\

\noindent\textbf{Ablations.}
We create several ablations of our approach \approachNameShort
by using the reading component and answering module in conjunction and alternatively. (i) \textbf{I+Q:} This ablation is state-of-the-art for VQA 2.0 and doesn't use any kind of OCR features; we provide results on Pythia v0.3 and BAN \cite{kim2018bilinear} in Tab.~\ref{tab:vqaaccuracy}; (ii) \textbf{Pythia+O:} Pythia with OCR features as input but no copy module or dynamic answer space; (iii)~\textbf{Pythia+O+C:} (ii) with the copy mechanism but no fixed answer space \ie the model can only predict from the OCR tokens. Abbreviation \textbf{C} is used when we add the copy module and dynamic answer space to a model.

Our full model corresponds to \textbf{LoRRA} attached to Pythia. We also compare \textbf{Pythia+LoRRA}  with small answer space \textbf{(SA)} to a version with large answer space \textbf{(LA)}. We also provide results on \textbf{LoRRA} attached to BAN~\cite{kim2018bilinear}.
\\

\noindent\textbf{Experimental Setup.}
\label{subsec:setup}
We develop our model in PyTorch \cite{paszke2017automatic}. We use AdaMax optimizer \cite{kingma2014adam} to perform back-propagation \cite{lecun1989backpropagation}. We predict logits and train using binary cross-entropy loss. We train all of our models for 24000 iterations with a batch size of 128 on 8 GPUs. We set the maximum question length to 14 and maximum number of OCR tokens to 50. We pad rest of the sequence if it is less than the maximum length. 
We use a learning rate of 5e-2 for all layers except the $fc7$ layers used for fine-tuning which are trained with 5e-3. We uniformly decrease the learning rate to 5e-4 after 14k iterations. We calculate val accuracy using VQA accuracy metric~\cite{balanced_vqa_v2} at every 1000th iteration and use the model with the best validation accuracy to calculate the test accuracy. All validation accuracies are averaged over 5 runs with different seeds.
\\

\noindent\textbf{Results.}
\label{subsec:results}
Tab.~\ref{tab:results} shows accuracies on both heuristics (left) and trained baselines and models (right). Despite collecting open-ended answers from annotators, we find that human accuracy is 85.01\%, consistent with that on VQA 2.0~\cite{balanced_vqa_v2} and VizWiz~\cite{gurari2018vizwiz}. While the OCR system we used is not perfect, the upper-bound on the validation set that one can achieve by correctly predicting the answer using these OCR tokens is 37.12\%. This is higher than our best model, suggesting room for improvement to reason about the OCR tokens. 
\textbf{LA UB} is quite high as they contain most commonly occurring questions.
This accuracy on VQA 2.0 validation set with 3129 most common answers is 88.9\% which suggests uniqueness of answers in \datasetName and limits of a fixed answer space.
The difference between \approachNameShort and \textbf{LA+OCR UB} of 41\% represents the room for improvement in modelling with current OCR tokens and \textbf{LA}.
Majority answer (``yes") gets only 4.48\% on test set.
Random baselines, even the weighted one, are rarely correct.
\textbf{Random OCR} token selection and maximum occurring OCR token selection (\textbf{OCR Max}) yields better accuracies compared to other heuristics baselines.
Question only \textbf{(Q)} and Image only \textbf{(I)} baseline get 8.09\% and 6.29\% validation accuracies, respectively, which shows that the dataset does not have significant biases \wrt images and questions. 
\textbf{I+Q} models - Pythia v0.3~\cite{singhpythia} and BAN~\cite{kim2018bilinear}, which
 are state-of-the-art on VQA 2.0 and VizWiz
only achieve 13.04\% and 12.3\% validation accuracy on \datasetName, respectively. 
This demonstrates the inability of current VQA models to read and reason about text in images. A  jump in accuracy to 18.35\% is observed by feeding OCR tokens \textbf{(Pythia+O)} into the model; this supports the hypothesis that OCR tokens do help in predicting correct answers. Validation accuracy of 20.06 achieved by \textbf{Pythia+O+C} by only 
predicting answers from OCR tokens, further bolsters OCR importance as it is quite high compared to our Pythia v0.3 \cite{singhpythia}. 

Our \textbf{LoRRA (LA)} with Pythia model outperforms all of the ablations. Finally, a slight modification which allows the model to predict from the OCR tokens more often by changing the fixed answer space \textbf{LA} to \textbf{SA} further improves performance. Validation accuracy for BAN~\cite{kim2018bilinear} also improves to 18.41\% by adding LoRRA. This suggests that LoRRA can help state-of-the-art VQA models to perform better on \datasetName.

While  \approachNameShort can reach up to 26.56\% accuracy on the \datasetName's validation set, there is a large gap to human performance of 85.01\% and LA+OCR UB of 67.56\%.

Interestingly, when adding LoRRA to Pythia it improves accuracy from 68.71 to 69.21 on  VQA 2.0 \cite{goyal2017accurate} (see Tab.~\ref{tab:vqaaccuracy}), indicating the  ability of our model to also exploit reading and reasoning in this more general VQA benchmark.

\section{Conclusion}

We explore a specific skill in Visual Question Answering that is important for the 
applications involving aiding 
visually impaired users -- answering 
questions about everyday images that involve reading and reasoning about text in these images. We find that
existing datasets do not support a systematic exploration of the research efforts towards this goal.
To this end, we introduce the 
\datasetName dataset which contains questions which can only be answered by reading and reasoning about text in images.  
We also introduce \approachName, a novel model architecture for answering questions based on text in images. \approachNameShort reads the text in images, reasons about it based on the provided question, and predicts an answer from a fixed vocabulary or the text found in the image. \approachNameShort is agnostic to the specifics of the underlying OCR and VQA modules.
\approachNameShort significantly outperforms the current state-of-the-art VQA models on \datasetName.
Our OCR model, while mature, still fails at detecting text that is rotated,
a bit unstructured (e.g., a scribble) or partially occluded. We believe 
\datasetName will encourage research both on improving text detection and recognition in unconstrained environments as well as on enabling the VQA models to read and reason about text in images.

{\small
\bibliographystyle{ieee}
\bibliography{egbib}

\begin{thebibliography}{10}\itemsep=-1pt

\bibitem{anderson2017bottom}
Peter Anderson, Xiaodong He, Chris Buehler, Damien Teney, Mark Johnson, Stephen
  Gould, and Lei Zhang.
\newblock Bottom-up and top-down attention for image captioning and visual
  question answering.
\newblock In {\em Computer Vision and Pat-tern Recognition (CVPR)}, 2018.

\bibitem{andreas2016neural}
Jacob Andreas, Marcus Rohrbach, Trevor Darrell, and Dan Klein.
\newblock Neural module networks.
\newblock In {\em Computer Vision and Pat-tern Recognition (CVPR)}, 2016.

\bibitem{antol2015vqa}
Stanislaw Antol, Aishwarya Agrawal, Jiasen Lu, Margaret Mitchell, Dhruv Batra,
  C Lawrence~Zitnick, and Devi Parikh.
\newblock Vqa: Visual question answering.
\newblock In {\em ICCV}, 2015.

\bibitem{bahdanau2014neural}
Dzmitry Bahdanau, Kyunghyun Cho, and Yoshua Bengio.
\newblock Neural machine translation by jointly learning to align and
  translate.
\newblock In {\em International Conference on Learning Representations (ICLR)},
  2015.

\bibitem{bigham2010vizwiz}
Jeffrey~P Bigham, Chandrika Jayant, Hanjie Ji, Greg Little, Andrew Miller,
  Robert~C Miller, Robin Miller, Aubrey Tatarowicz, Brandyn White, Samual
  White, et~al.
\newblock Vizwiz: nearly real-time answers to visual questions.
\newblock In {\em Proceedings of the 23nd annual ACM symposium on User
  interface software and technology}, pages 333--342. ACM, 2010.

\bibitem{borisyuk2018rosetta}
Fedor Borisyuk, Albert Gordo, and Viswanath Sivakumar.
\newblock Rosetta: Large scale system for text detection and recognition in
  images.
\newblock In {\em Proceedings of the 24th ACM SIGKDD International Conference
  on Knowledge Discovery \& Data Mining}, pages 71--79. ACM, 2018.

\bibitem{fukui2016multimodal}
Akira Fukui, Dong~Huk Park, Daylen Yang, Anna Rohrbach, Trevor Darrell, and
  Marcus Rohrbach.
\newblock Multimodal compact bilinear pooling for visual question answering and
  visual grounding.
\newblock In {\em EMNLP}, 2016.

\bibitem{Detectron2018}
Ross Girshick, Ilija Radosavovic, Georgia Gkioxari, Piotr Doll\'{a}r, and
  Kaiming He.
\newblock Detectron.
\newblock \url{https://github.com/facebookresearch/detectron}, 2018.

\bibitem{goyal2017accurate}
Priya Goyal, Piotr Doll{\'a}r, Ross Girshick, Pieter Noordhuis, Lukasz
  Wesolowski, Aapo Kyrola, Andrew Tulloch, Yangqing Jia, and Kaiming He.
\newblock Accurate, large minibatch sgd: training imagenet in 1 hour.
\newblock {\em arXiv preprint arXiv:1706.02677}, 2017.

\bibitem{balanced_vqa_v2}
Yash Goyal, Tejas Khot, Douglas Summers{-}Stay, Dhruv Batra, and Devi Parikh.
\newblock Making the {V} in {VQA} matter: Elevating the role of image
  understanding in {V}isual {Q}uestion {A}nswering.
\newblock In {\em Conference on Computer Vision and Pattern Recognition
  (CVPR)}, 2017.

\bibitem{gu2016incorporating}
J Gu, Z Lu, H Li, and VOK Li.
\newblock Incorporating copying mechanism in sequence-to-sequence learning.
\newblock In {\em Annual Meeting of the Association for Computational
  Linguistics (ACL), 2016}. Association for Computational Linguistics., 2016.

\bibitem{gulcehre2016pointing}
Caglar Gulcehre, Sungjin Ahn, Ramesh Nallapati, Bowen Zhou, and Yoshua Bengio.
\newblock Pointing the unknown words.
\newblock In {\em ACL}, 2016.

\bibitem{gurari2018vizwiz}
Danna Gurari, Qing Li, Abigale~J Stangl, Anhong Guo, Chi Lin, Kristen Grauman,
  Jiebo Luo, and Jeffrey~P Bigham.
\newblock Vizwiz grand challenge: Answering visual questions from blind people.
\newblock In {\em Conference on Computer Vision and PatternRecognition (CVPR)},
  2017.

\bibitem{he2016deep}
Kaiming He, Xiangyu Zhang, Shaoqing Ren, and Jian Sun.
\newblock Deep residual learning for image recognition.
\newblock In {\em Proceedings of the IEEE conference on computer vision and
  pattern recognition}, pages 770--778, 2016.

\bibitem{hochreiter1997long}
Sepp Hochreiter and J{\"u}rgen Schmidhuber.
\newblock Long short-term memory.
\newblock {\em Neural computation}, 9(8):1735--1780, 1997.

\bibitem{jiang2017memexqa}
Lu Jiang, Junwei Liang, Liangliang Cao, Yannis Kalantidis, Sachin Farfade, and
  Alexander~G Hauptmann.
\newblock Memexqa: Visual memex question answering.
\newblock {\em arXiv:1708.01336}, 2017.

\bibitem{jiang2018pythia}
Yu Jiang, Vivek Natarajan, Xinlei Chen, Marcus Rohrbach, Dhruv Batra, and Devi
  Parikh.
\newblock Pythia v0. 1: the winning entry to the vqa challenge 2018.
\newblock {\em arXiv preprint arXiv:1807.09956}, 2018.

\bibitem{johnson2017clevr}
Justin Johnson, Bharath Hariharan, Laurens van~der Maaten, Li Fei-Fei,
  C~Lawrence Zitnick, and Ross Girshick.
\newblock Clevr: A diagnostic dataset for compositional language and elementary
  visual reasoning.
\newblock In {\em Computer Vision and Pattern Recognition (CVPR), 2017 IEEE
  Conference on}, pages 1988--1997. IEEE, 2017.

\bibitem{joulin2016bag}
Armand Joulin, Edouard Grave, Piotr Bojanowski, and Tomas Mikolov.
\newblock Bag of tricks for efficient text classification.
\newblock In {\em European Chapter of the Association for Computational
  Linguistics}, 2017.

\bibitem{kafle2018dvqa}
Kushal Kafle, Scott Cohen, Brian Price, and Christopher Kanan.
\newblock Dvqa: Understanding data visualizations via question answering.
\newblock In {\em Proceedings of the IEEE Conference on Computer Vision and
  Pattern Recognition}, pages 5648--5656, 2018.

\bibitem{kahou2017figureqa}
Samira~Ebrahimi Kahou, Vincent Michalski, Adam Atkinson, Akos Kadar, Adam
  Trischler, and Yoshua Bengio.
\newblock Figureqa: An annotated figure dataset for visual reasoning.
\newblock In {\em ICLR workshop track}, 2018.

\bibitem{karatzas2015icdar}
Dimosthenis Karatzas, Lluis Gomez-Bigorda, Anguelos Nicolaou, Suman Ghosh,
  Andrew Bagdanov, Masakazu Iwamura, Jiri Matas, Lukas Neumann,
  Vijay~Ramaseshan Chandrasekhar, Shijian Lu, et~al.
\newblock Icdar 2015 competition on robust reading.
\newblock In {\em Document Analysis and Recognition (ICDAR), 2015 13th
  International Conference on}, pages 1156--1160. IEEE, 2015.

\bibitem{kembhavi2016diagram}
Aniruddha Kembhavi, Mike Salvato, Eric Kolve, Minjoon Seo, Hannaneh Hajishirzi,
  and Ali Farhadi.
\newblock A diagram is worth a dozen images.
\newblock In {\em European Conference on Computer Vision}, pages 235--251.
  Springer, 2016.

\bibitem{kembhavi2017you}
Aniruddha Kembhavi, Min~Joon Seo, Dustin Schwenk, Jonghyun Choi, Ali Farhadi,
  and Hannaneh Hajishirzi.
\newblock Are you smarter than a sixth grader? textbook question answering for
  multimodal machine comprehension.
\newblock In {\em Computer Vision and Pat-tern Recognition (CVPR)}, volume~2,
  page~3, 2017.

\bibitem{kim2018bilinear}
Jin-Hwa Kim, Jaehyun Jun, and Byoung-Tak Zhang.
\newblock Bilinear attention networks.
\newblock In {\em Neural Information Processing Systems}, 2018.

\bibitem{kingma2014adam}
Diederik~P Kingma and Jimmy Ba.
\newblock Adam: A method for stochastic optimization.
\newblock In {\em International Conference on Learning Representations (ICLR)},
  2015.

\bibitem{krasin2016openimages}
Ivan Krasin, Tom Duerig, Neil Alldrin, Andreas Veit, Sami Abu-El-Haija, Serge
  Belongie, David Cai, Zheyun Feng, Vittorio Ferrari, Victor Gomes, et~al.
\newblock Openimages: A public dataset for large-scale multi-label and
  multi-class image classification.
\newblock {\em Dataset available from https://github. com/openimages}, 2(6):7,
  2016.

\bibitem{krishna2017visual}
Ranjay Krishna, Yuke Zhu, Oliver Groth, Justin Johnson, Kenji Hata, Joshua
  Kravitz, Stephanie Chen, Yannis Kalantidis, Li-Jia Li, David~A Shamma, et~al.
\newblock Visual genome: Connecting language and vision using crowdsourced
  dense image annotations.
\newblock {\em IJCV}, 2017.

\bibitem{lecun1989backpropagation}
Yann LeCun, Bernhard Boser, John~S Denker, Donnie Henderson, Richard~E Howard,
  Wayne Hubbard, and Lawrence~D Jackel.
\newblock Backpropagation applied to handwritten zip code recognition.
\newblock {\em Neural computation}, 1(4):541--551, 1989.

\bibitem{lu2016hierarchical}
Jiasen Lu, Jianwei Yang, Dhruv Batra, and Devi Parikh.
\newblock Hierarchical question-image co-attention for visual question
  answering.
\newblock In {\em Advances In Neural Information Processing Systems}, pages
  289--297, 2016.

\bibitem{malinowski2014multi}
Mateusz Malinowski and Mario Fritz.
\newblock A multi-world approach to question answering about real-world scenes
  based on uncertain input.
\newblock In {\em Advances in neural information processing systems}, pages
  1682--1690, 2014.

\bibitem{merity2016pointer}
Stephen Merity, Caiming Xiong, James Bradbury, and Richard Socher.
\newblock Pointer sentinel mixture models.
\newblock In {\em International Conference on Learning Representations (ICLR)},
  2017.

\bibitem{mishra2012scene}
Anand Mishra, Karteek Alahari, and CV Jawahar.
\newblock Scene text recognition using higher order language priors.
\newblock In {\em BMVC-British Machine Vision Conference}. BMVA, 2012.

\bibitem{nallapati2016abstractive}
Ramesh Nallapati, Bowen Zhou, Caglar Gulcehre, Bing Xiang, et~al.
\newblock Abstractive text summarization using sequence-to-sequence rnns and
  beyond.
\newblock In {\em The SIGNLL Conference on Computational Natural Language
  Learning (CoNLL)}, 2016.

\bibitem{paszke2017automatic}
Adam Paszke, Sam Gross, Soumith Chintala, Gregory Chanan, Edward Yang, Zachary
  DeVito, Zeming Lin, Alban Desmaison, Luca Antiga, and Adam Lerer.
\newblock Automatic differentiation in pytorch.
\newblock {\em NIPS AutoDiff Workshop}, 2017.

\bibitem{pennington2014glove}
Jeffrey Pennington, Richard Socher, and Christopher Manning.
\newblock Glove: Global vectors for word representation.
\newblock In {\em EMNLP}, 2014.

\bibitem{raghu2018hierarchical}
Dinesh Raghu, Nikhil Gupta, et~al.
\newblock Hierarchical pointer memory network for task oriented dialogue.
\newblock {\em arXiv preprint arXiv:1805.01216}, 2018.

\bibitem{ren2015exploring}
Mengye Ren, Ryan Kiros, and Richard Zemel.
\newblock Exploring models and data for image question answering.
\newblock In {\em Advances in neural information processing systems}, pages
  2953--2961, 2015.

\bibitem{see2017get}
Abigail See, Peter~J Liu, and Christopher~D Manning.
\newblock Get to the point: Summarization with pointer-generator networks.
\newblock In {\em Association for Computational Linguistics}, 2017.

\bibitem{singhpythia}
Amanpreet Singh, Vivek Natarajan, Yu Jiang, Xinlei Chen, Meet Shah, Marcus
  Rohrbach, Dhruv Batra, and Devi Parikh.
\newblock Pythia-a platform for vision \& language research.
\newblock {\em SysML Workshop, NeurIPS 2019}, 2018.

\bibitem{smith2007overview}
Ray Smith.
\newblock An overview of the tesseract ocr engine.
\newblock In {\em Document Analysis and Recognition, 2007. ICDAR 2007. Ninth
  International Conference on}, volume~2, pages 629--633. IEEE, 2007.

\bibitem{suhr2017corpus}
Alane Suhr, Mike Lewis, James Yeh, and Yoav Artzi.
\newblock A corpus of natural language for visual reasoning.
\newblock In {\em Proceedings of the 55th Annual Meeting of the Association for
  Computational Linguistics (Volume 2: Short Papers)}, volume~2, pages
  217--223, 2017.

\bibitem{veit2016coco}
Andreas Veit, Tomas Matera, Lukas Neumann, Jiri Matas, and Serge Belongie.
\newblock Coco-text: Dataset and benchmark for text detection and recognition
  in natural images.
\newblock {\em arXiv preprint arXiv:1601.07140}, 2016.

\bibitem{wang2010word}
Kai Wang and Serge Belongie.
\newblock Word spotting in the wild.
\newblock In {\em European Conference on Computer Vision}, pages 591--604.
  Springer, 2010.

\bibitem{wang2018fvqa}
Peng Wang, Qi Wu, Chunhua Shen, Anthony Dick, and Anton van~den Hengel.
\newblock Fvqa: Fact-based visual question answering.
\newblock {\em IEEE transactions on pattern analysis and machine intelligence},
  2018.

\bibitem{xiong2016dynamic}
Caiming Xiong, Victor Zhong, and Richard Socher.
\newblock Dynamic coattention networks for question answering.
\newblock In {\em International Conference on Learning Representations (ICLR)},
  2016.

\bibitem{xu2016ask}
Huijuan Xu and Kate Saenko.
\newblock Ask, attend and answer: Exploring question-guided spatial attention
  for visual question answering.
\newblock In {\em European Conference on Computer Vision}, pages 451--466.
  Springer, 2016.

\bibitem{yang2016stacked}
Zichao Yang, Xiaodong He, Jianfeng Gao, Li Deng, and Alex Smola.
\newblock Stacked attention networks for image question answering.
\newblock In {\em Computer Vision and Pat-tern Recognition (CVPR)}, 2016.

\bibitem{yu2018beyond}
Zhou Yu, Jun Yu, Chenchao Xiang, Jianping Fan, and Dacheng Tao.
\newblock Beyond bilinear: Generalized multimodal factorized high-order pooling
  for visual question answering.
\newblock {\em IEEE Transactions on Neural Networks and Learning Systems},
  2018.

\bibitem{zhang2018counting}
Yan Zhang, Jonathon Hare, and Adam Pr{\"u}gel-Bennett.
\newblock Learning to count objects in natural images for visual question
  answering.
\newblock In {\em International Conference on Learning Representations (ICLR)},
  2018.

\bibitem{zhu2016visual7w}
Yuke Zhu, Oliver Groth, Michael Bernstein, and Li Fei-Fei.
\newblock Visual7w: Grounded question answering in images.
\newblock In {\em Proceedings of the IEEE Conference on Computer Vision and
  Pattern Recognition}, pages 4995--5004, 2016.

\end{thebibliography}
}
\newpage
\appendix
\section{OCR and Answer Space Analysis}
\label{appendx:analysis}
We perform the following analysis on \datasetName's validation set. We find that 44.9\% of \approachNameShort's predicted answers are from OCR tokens (i.e., using the copy mechanism). The remaining 55.1\% of predicted answers are from the pre-determined (short) answer vocabulary (SA). This shows that our approach does in fact rely heavily on what it reads in the image, and relies on its copy mechanism to generalize and produce answers that have never been seen or are rare in the training data.
While predicting answers from OCR tokens, the model gets the entire answer string correct 27\% of the time, and partially correct (i.e., matches one word in answer) 11\% of the time. The percentage of partially correct answers indicates the possibility of getting better results by using n-grams of OCR tokens or spelling correction for improving OCR predictions. 
When predicting from the answer space, the model gets the answer correct 22.4\% of the time. 

We find that 30.6\% of questions have their answers in OCR tokens. For these questions, \approachNameShort chooses to predict from OCR tokens 68\% of the times and answers 57.5\% of these correct. Similarly, 48\% of questions have their answers in SA. For these questions, \approachNameShort chooses to predict from LA 66.75\%  of the times and gets 38\% of these correct.

81\% of the questions in TextVQA's validation set have images with 2 or more OCR tokens. Among these 4,645 questions, \approachNameShort chooses to copy from OCR tokens 49.7\% of the time and gets 24.3\% of these correct. This suggests that \approachNameShort doesn't randomly copy OCR token from a list of available tokens.

\section{\datasetName Examples and \approachNameShort Predictions}

In Fig.~\ref{fig:attention_total}, we show representative examples from our \datasetName dataset along with the predictions from Pythia+\approachNameShort.
Each example shows the ground truth answer, the predictions from \approachNameShort, whether the answer prediction was from OCR tokens or the pre-determined answer space, and attention weights for each of the OCR tokens. 
The examples indicate the following points:
\begin{itemize}
    \item The model is able to successfully answer questions about times, dates, brands, cities and places, and is often able to correctly spell them even if the OCR tokens had them misspelled (by picking an answer from the pre-determined answer space).
    See Fig. \ref{fig:attention_11} (short hand's hour), Fig. \ref{fig:attention_7} (birthday date),
    Fig. \ref{fig:attention_19} (picking out city ``london'' from the large amount of text), Fig. \ref{fig:attention_15} (samsung).
    \item The model is able to successfully answer questions involving colors and spatial reasoning.
    See Fig. \ref{fig:attention_5} (player on the right), Fig. \ref{fig:attention_6} (location of coin), Fig. \ref{fig:attention_3} (location of banner).
    See Fig. \ref{fig:attention_17} where the model needs to identify the correct sign based on multiple colors, or Fig. \ref{fig:attention_18} where the model needs to identify the correct sign in the red circle. Note that unlike most existing VQA models, the model does not seem to be biased toward ``stop'' for red signs. In Fig. \ref{fig:attention_1} the model needs to predict the correct number based on spatial reasoning between the two choices 7 and 14. 
    \item The model is also able to reason about basic sizes (less, greater, smallest) and shapes (circle). 
    See Fig. \ref{fig:attention_11} where the model needs to figure out which one is the shorter hand, or Fig. \ref{fig:attention_17} where the model needs to figure out which one is the lowest measurement among four.
    \item The model often predicts an answer from the answer space as informed by OCR tokens. See Fig. \ref{fig:attention_11} where the Pythia model (which doesn't use OCR) predicts 3, but our approach predicts 4 which is the correct answer. 
    \item The model often answers questions about cities with ``new york''. See Fig. \ref{fig:attention_10} where the model predicts New York instead of San Francisco. We have observed this bias in other city related questions as well.
    \item For yes/no questions, even though ``yes'' is the more common answer, the model does predict ``no'' frequently. See Fig. \ref{fig:attention_13}, Fig. \ref{fig:attention_12}.
    \item Sometimes when the answer is not in the answer space, but the partial answer is in OCR tokens, the model predicts the partial answer which is closest to the actual answer. See Fig. \ref{fig:attention_5} where the model predicts ``fly'' instead of ``fly emirates'', or Fig. \ref{fig:attention_7} where the model predicts only the birthday date ``19'', instead of ``may 19''. By construction our model can only copy a single OCR token, but our \datasetName dataset contains Q/A pairs which require copying multiple OCR tokens in the right order. Exploring this is an interesting direction for future work.
    \item The model sometimes gets seemingly simple questions wrong by predicting generic answers. See Fig. \ref{fig:attention_8} where the model can't predict ``embossed'' even though it is in the detected OCR tokens, or see Fig. \ref{fig:attention_2} where the model predicts most common letter ``g'' in the answer space instead of predicting based on ``a-2'' in the OCR tokens.
    \item The model has a strong dependency on the quality of OCR tokens produced. If the OCR module missed some text in the image, the model's output can be wrong. See Fig.~\ref{fig:attention_9} or Fig.~\ref{fig:attention_16} where the OCR tokens do not contain the ground truth answer or see Fig.~\ref{fig:attention_21} where the OCR system is unable to correctly read ``irig'' the second time.
\end{itemize}

\section{Interface Screenshots}

\begin{figure*}[h!]
    \centering
    \includegraphics[width=1\textwidth]{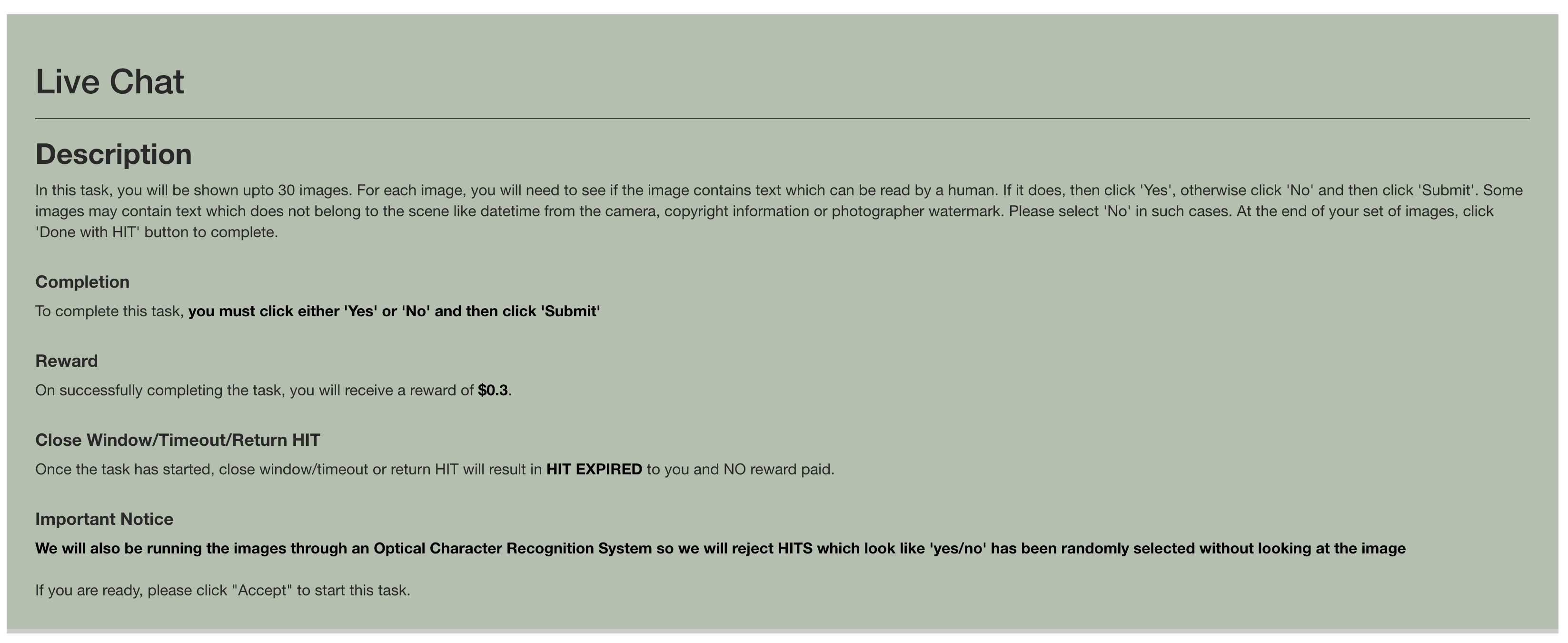}
    \caption{\textbf{Introduction page for our task.}
    }
    \label{fig:text_detection_entrance}
\end{figure*}

\begin{figure*}[h!]
    \centering
    \includegraphics[width=1\textwidth]{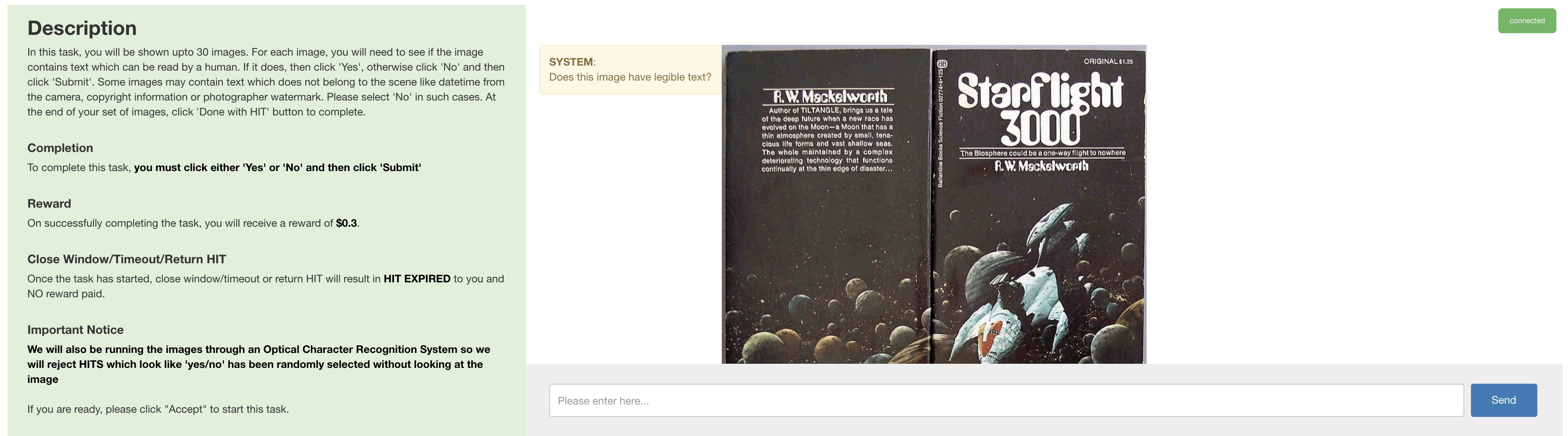}
    \caption{\textbf{Text detection task.} First stage of our data collection pipeline involves identifying and removing images without text.
    }
    \label{fig:text_detection_main}
\end{figure*}

\begin{figure*}[h!]
    \centering
    \includegraphics[width=1\textwidth]{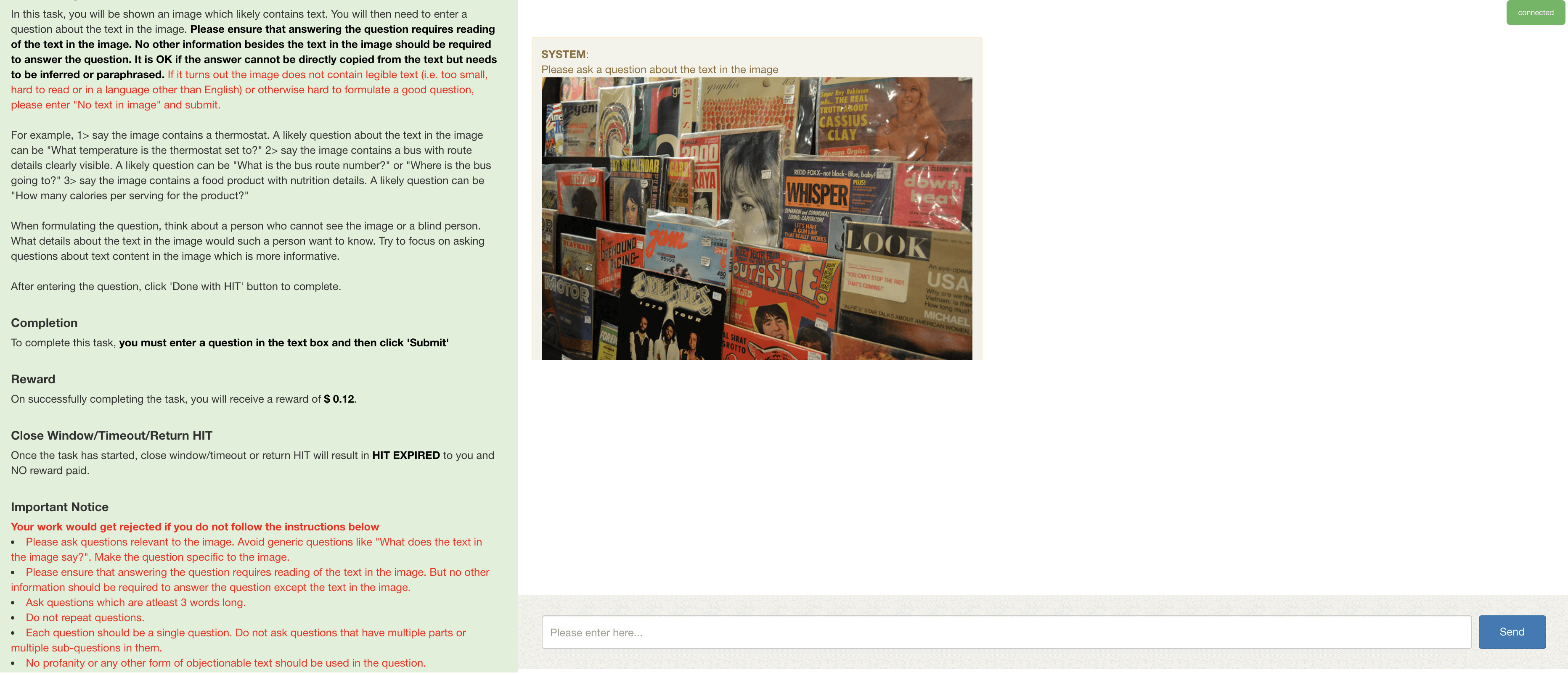}
    \caption{\textbf{Question task.} In the second stage, we ask workers to ask a question about an image whose answer requires reading text in the image. 
    We provide instructions and rules to ensure that we get high quality questions. 
    }
    \label{fig:question_main}
\end{figure*}

\begin{figure*}[h!]
    \centering
    \includegraphics[width=1\textwidth]{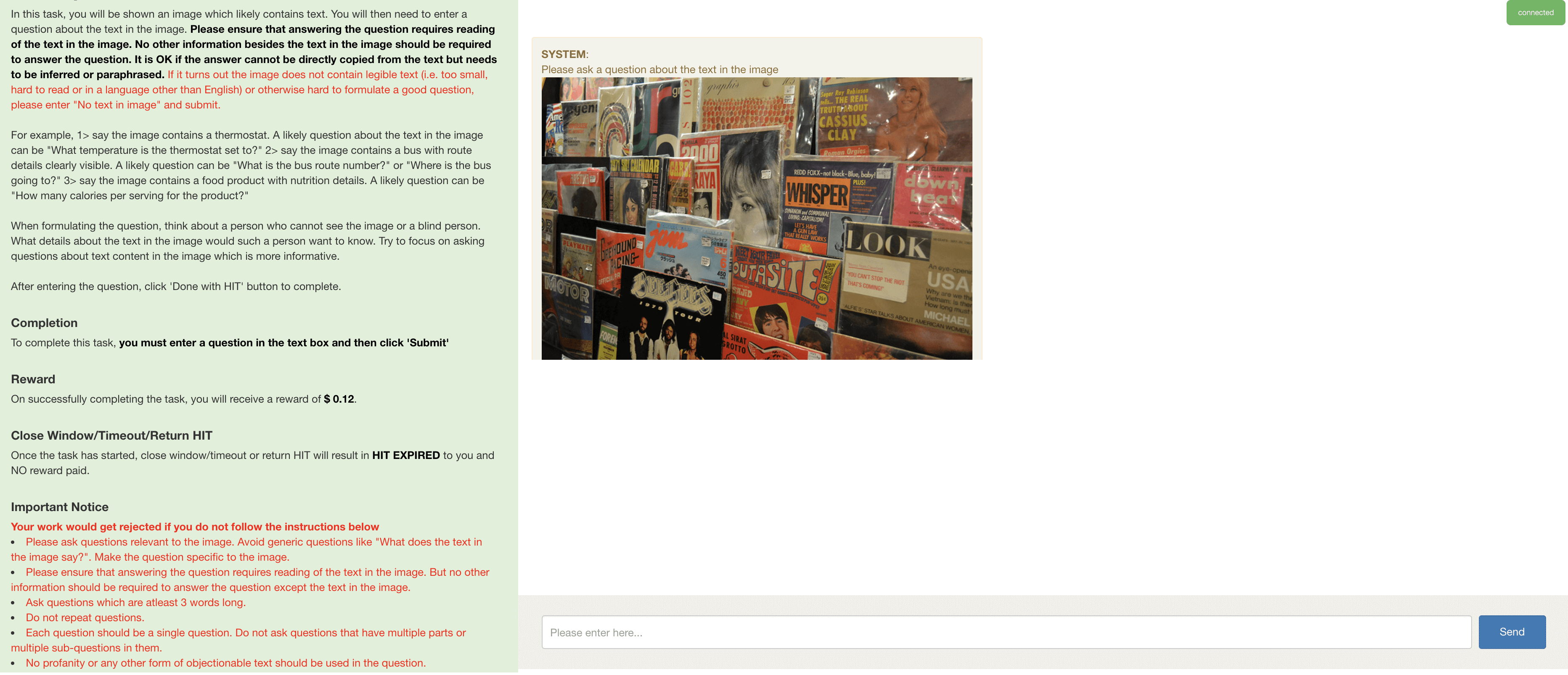}
    \caption{\textbf{Answer task.} In the third stage, we ask workers to answer a question about the image.
    }
    \label{fig:answer_main}
\end{figure*}

We show the three stages of the data collection pipeline in Fig.~\ref{fig:text_detection_entrance}, Fig.~\ref{fig:text_detection_main}, Fig.~\ref{fig:question_main} and Fig. \ref{fig:answer_main}. Fig. \ref{fig:text_detection_entrance} and Fig. \ref{fig:text_detection_main} shows the introduction and first stage of our pipeline which is used to identify and remove images without text in them. Fig \ref{fig:question_main} shows the second stage of our pipeline which is used to collect questions on images with text. Finally, the third stage interface is shown in Fig.~\ref{fig:answer_main} which is used to collect the answer for a question about an image.

\begin{figure*}
    \centering
    \begin{subfigure}[t]{0.32\textwidth}
        \includegraphics[width=1\linewidth]{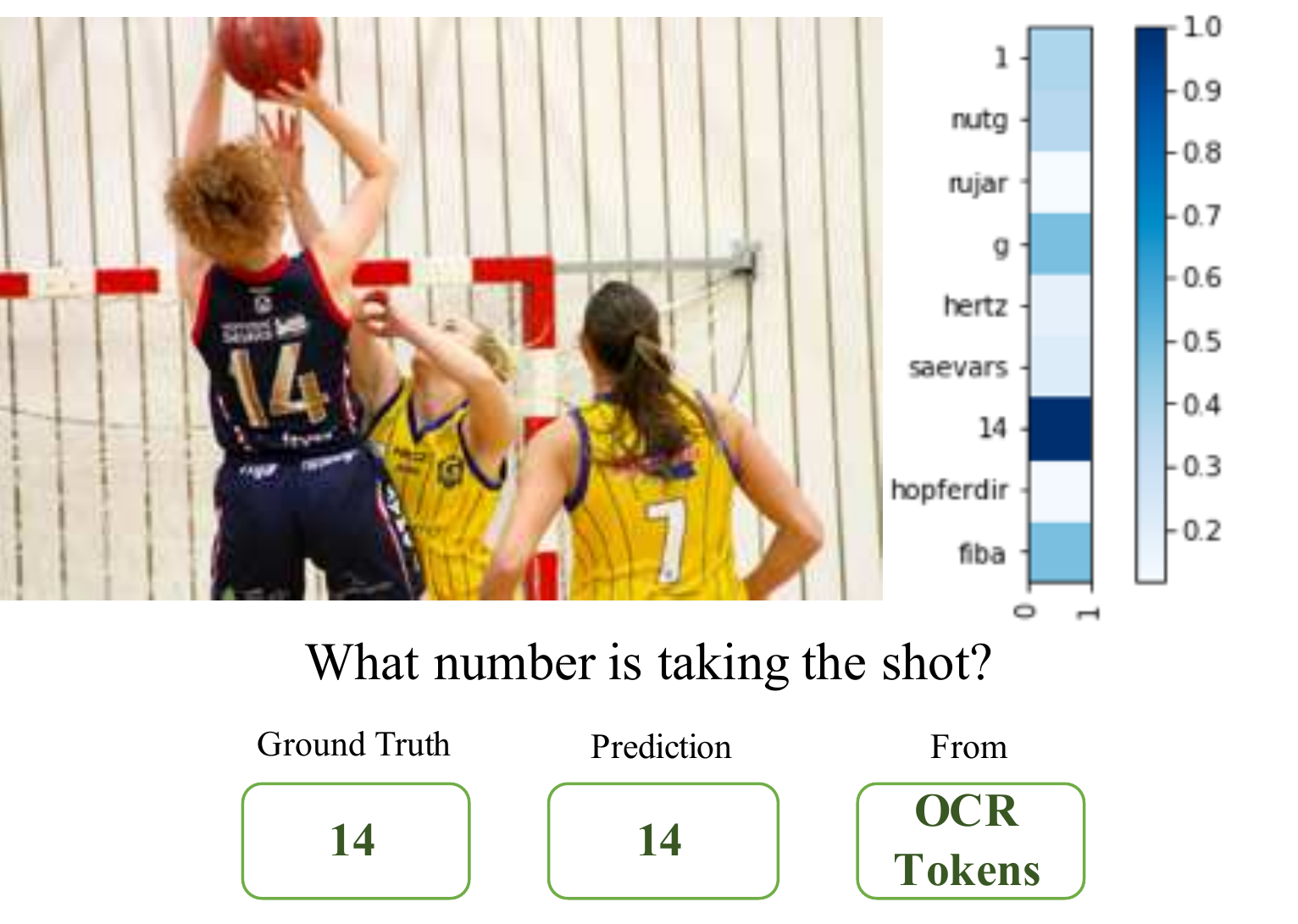}
        \caption{}
        \label{fig:attention_1}
    \end{subfigure}
    \begin{subfigure}[t]{0.32\textwidth}
        \includegraphics[width=1\linewidth]{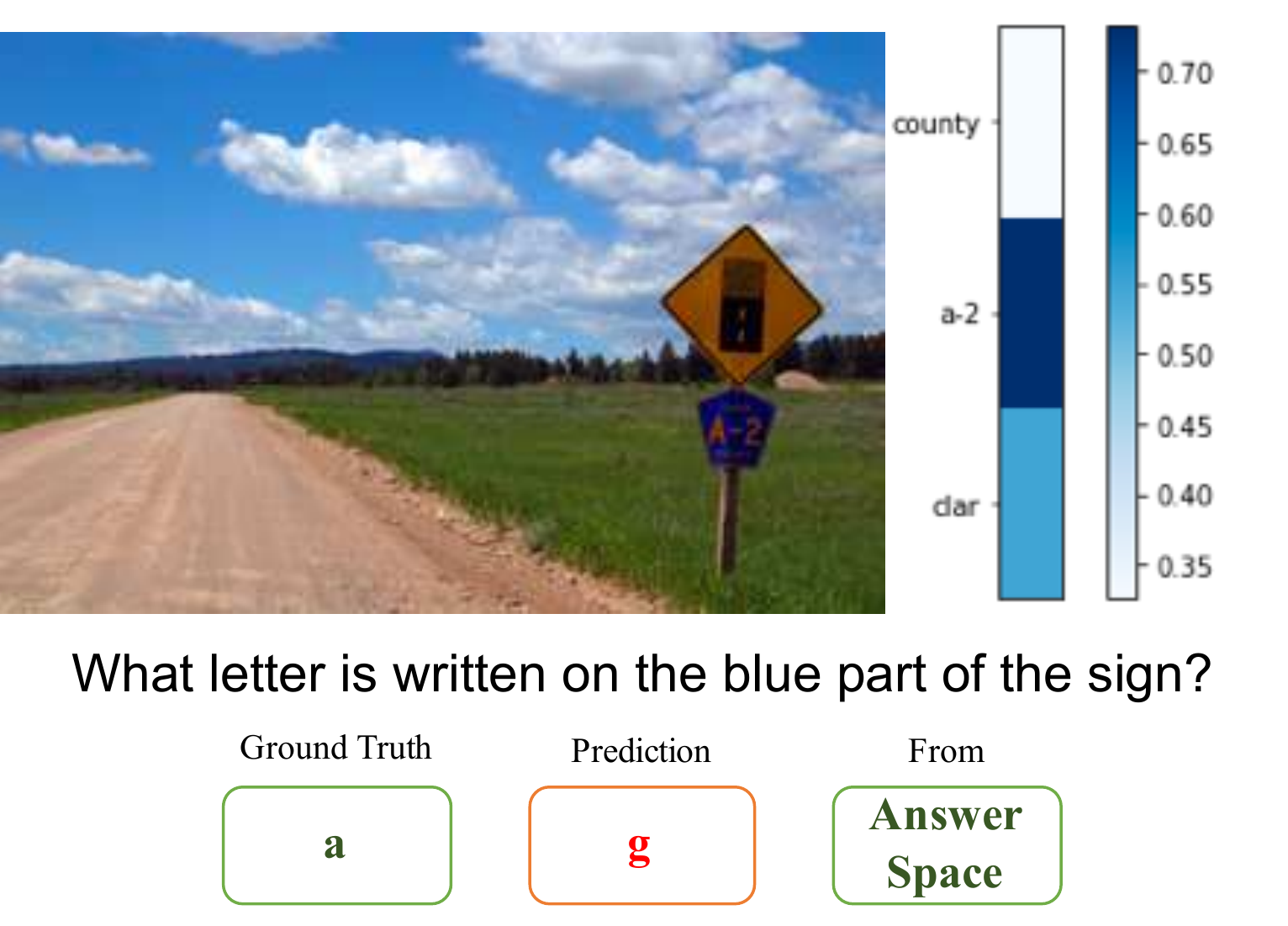}
        \caption{}
        \label{fig:attention_2}
    \end{subfigure}
        \begin{subfigure}[t]{0.32\textwidth}
        \includegraphics[width=1\linewidth]{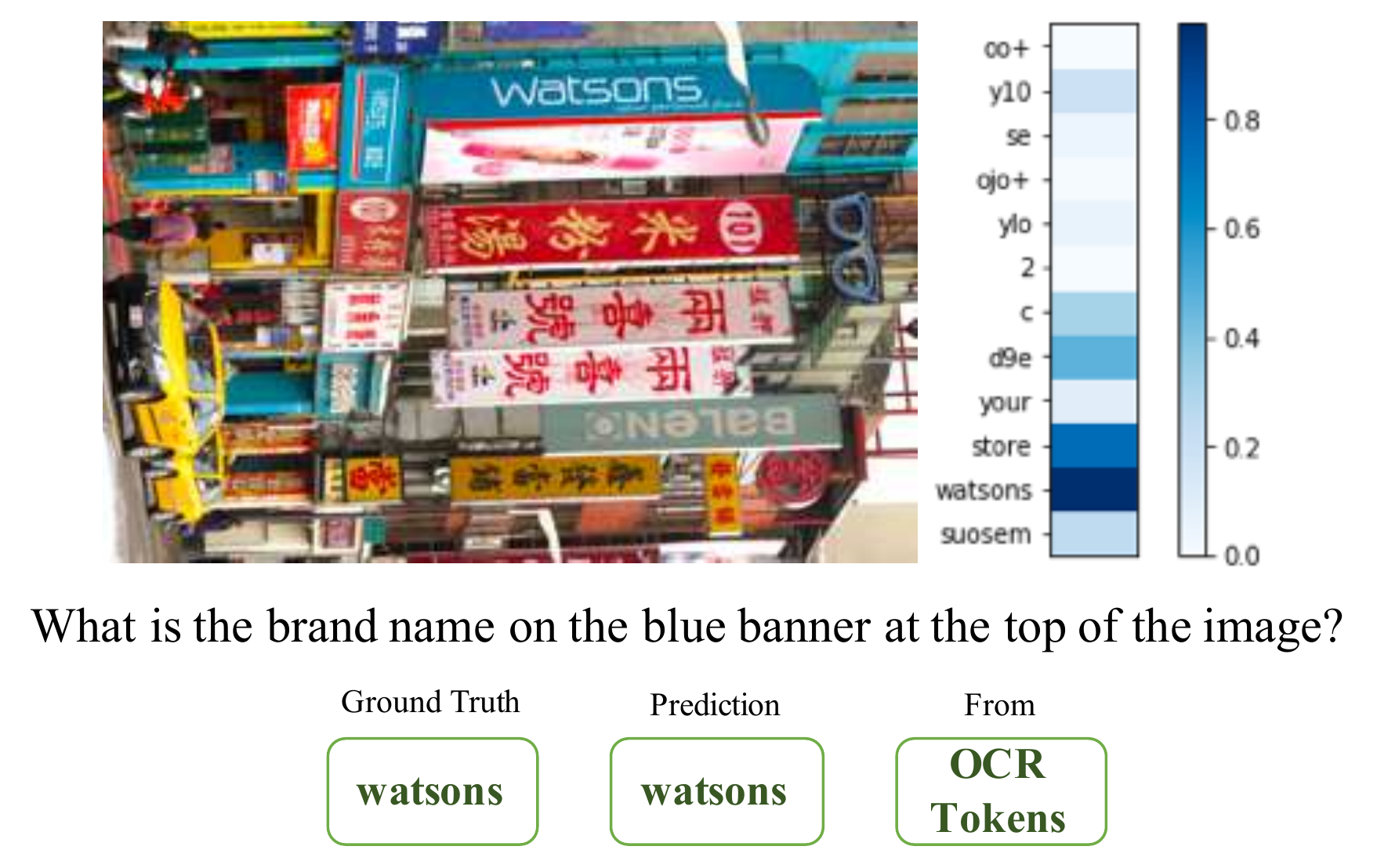}
        \caption{}
        \label{fig:attention_3}
    \end{subfigure}
    \begin{subfigure}[t]{0.32\textwidth}
        \includegraphics[width=1\linewidth]{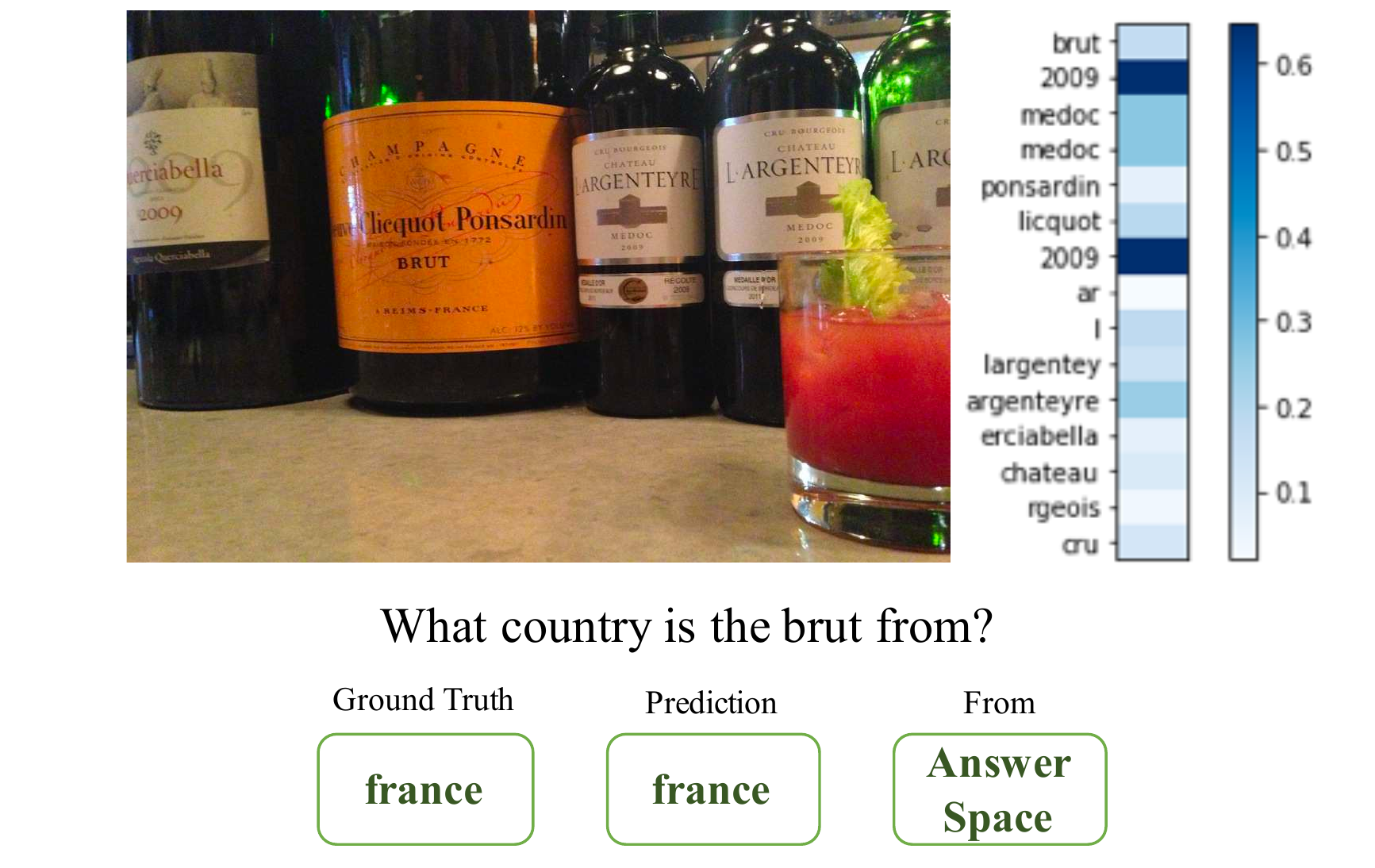}
        \caption{}
        \label{fig:attention_4}
    \end{subfigure}
    \begin{subfigure}[t]{0.32\textwidth}
        \includegraphics[width=1\linewidth]{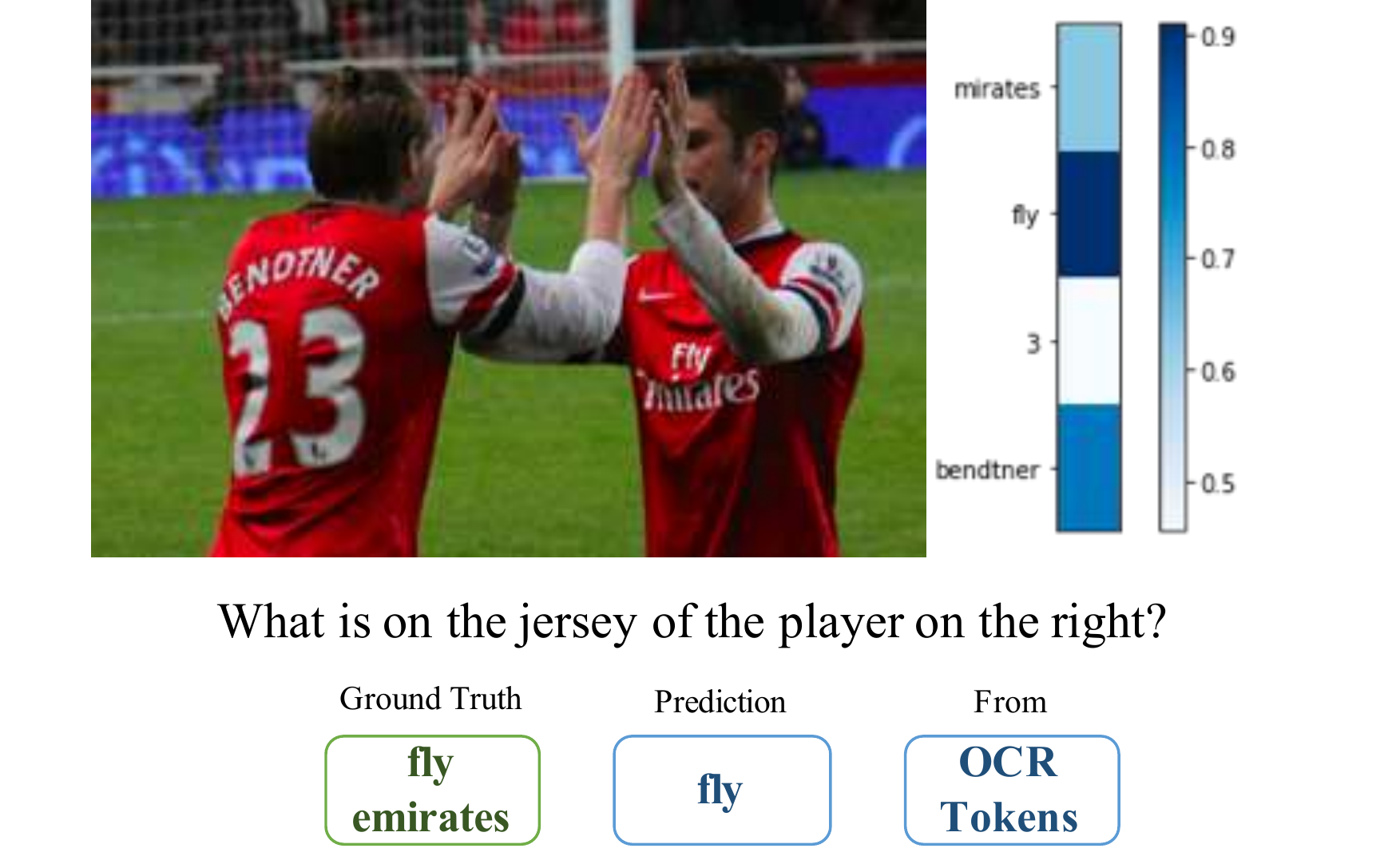}
        \caption{}
        \label{fig:attention_5}
    \end{subfigure}
    \begin{subfigure}[t]{0.32\textwidth}
        \includegraphics[width=1\linewidth]{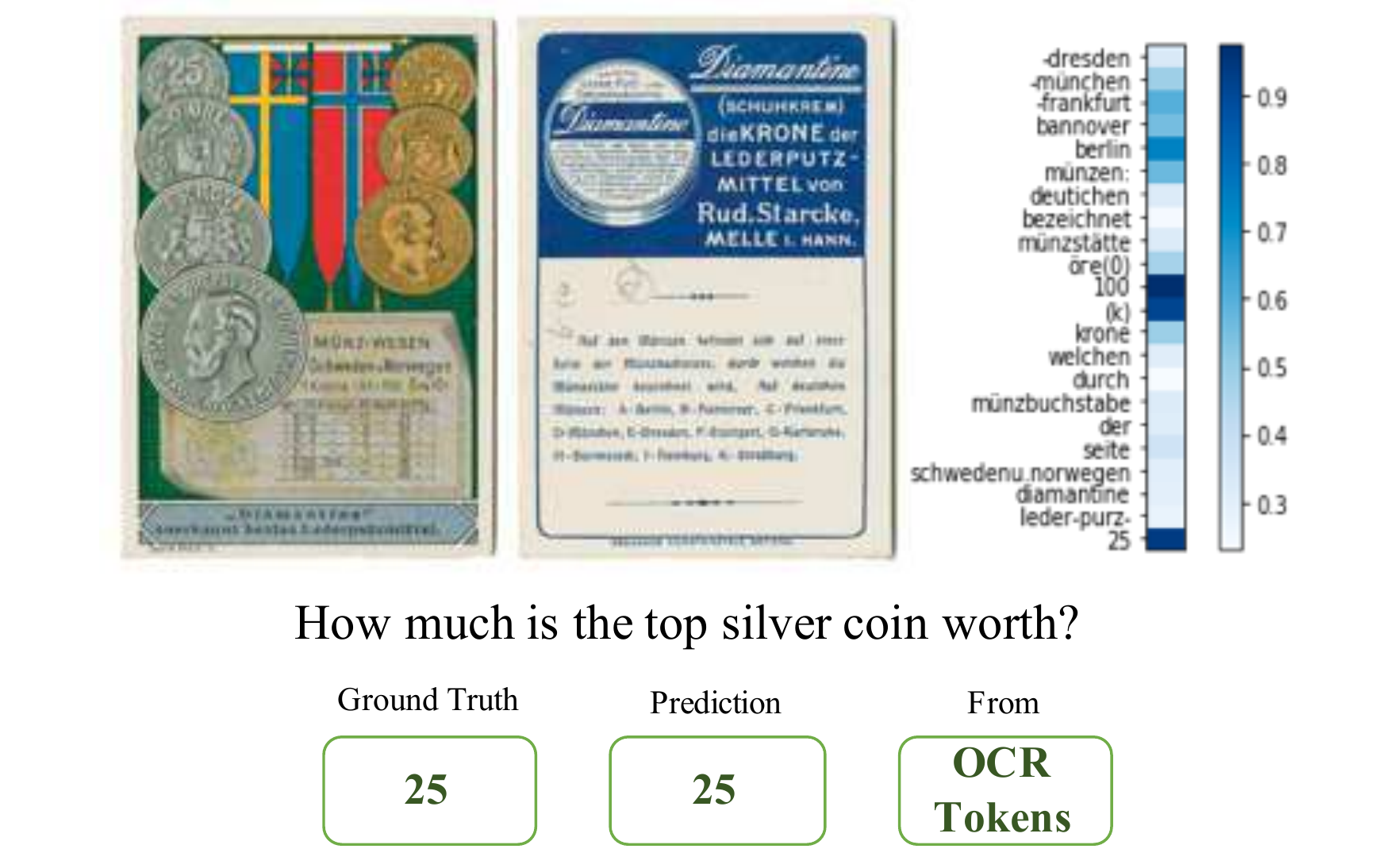}
        \caption{}
        \label{fig:attention_6}
    \end{subfigure}
    \begin{subfigure}[t]{0.32\textwidth}
        \includegraphics[width=1\linewidth]{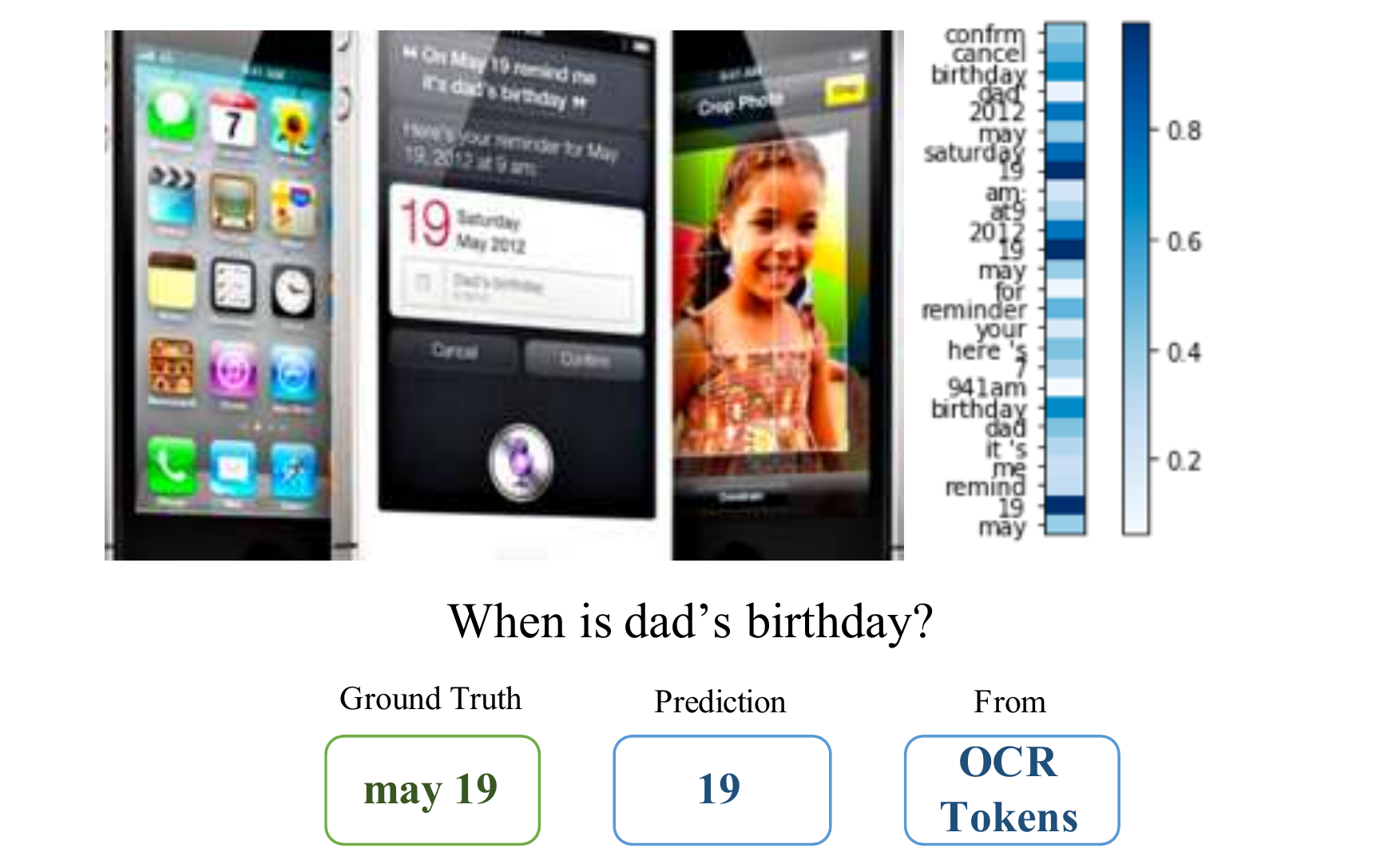}
        \caption{}
        \label{fig:attention_7}
    \end{subfigure}
        \begin{subfigure}[t]{0.32\textwidth}
        \includegraphics[width=1\linewidth]{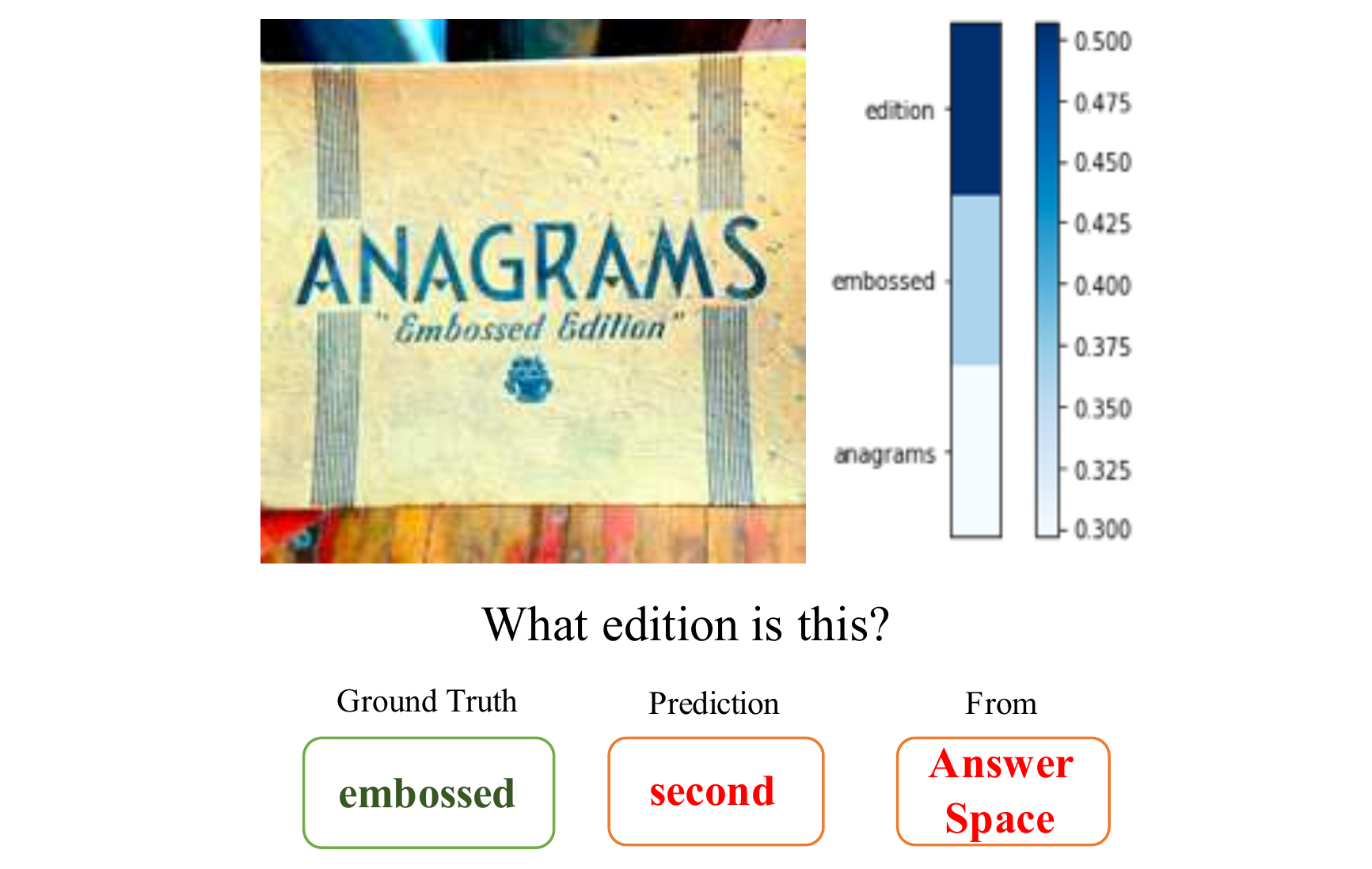}
        \caption{}
        \label{fig:attention_8}
    \end{subfigure}
    \begin{subfigure}[t]{0.32\textwidth}
        \includegraphics[width=1\linewidth]{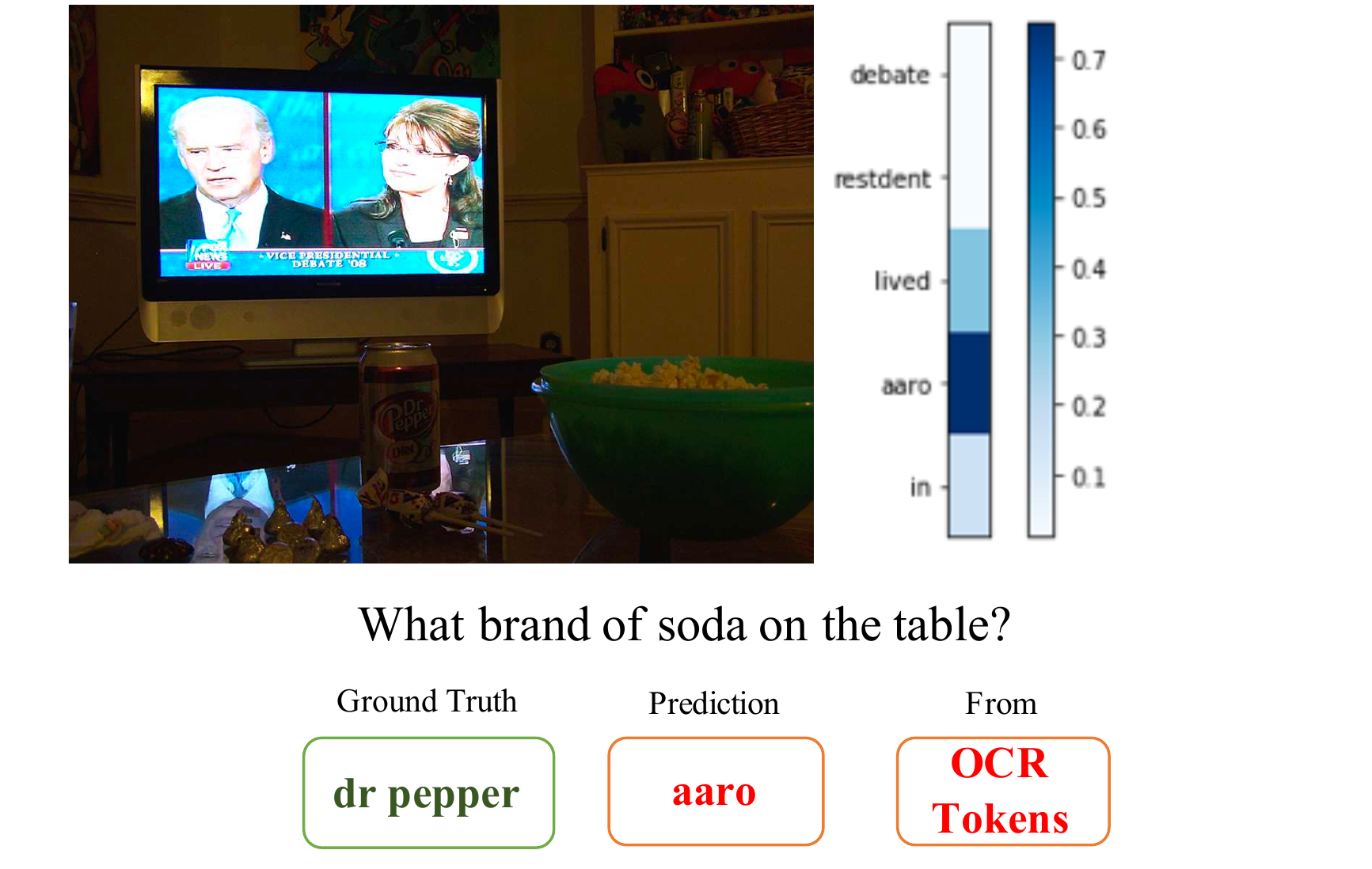}
        \caption{}
        \label{fig:attention_9}
    \end{subfigure}
    \begin{subfigure}[t]{0.32\textwidth}
        \includegraphics[width=1\linewidth]{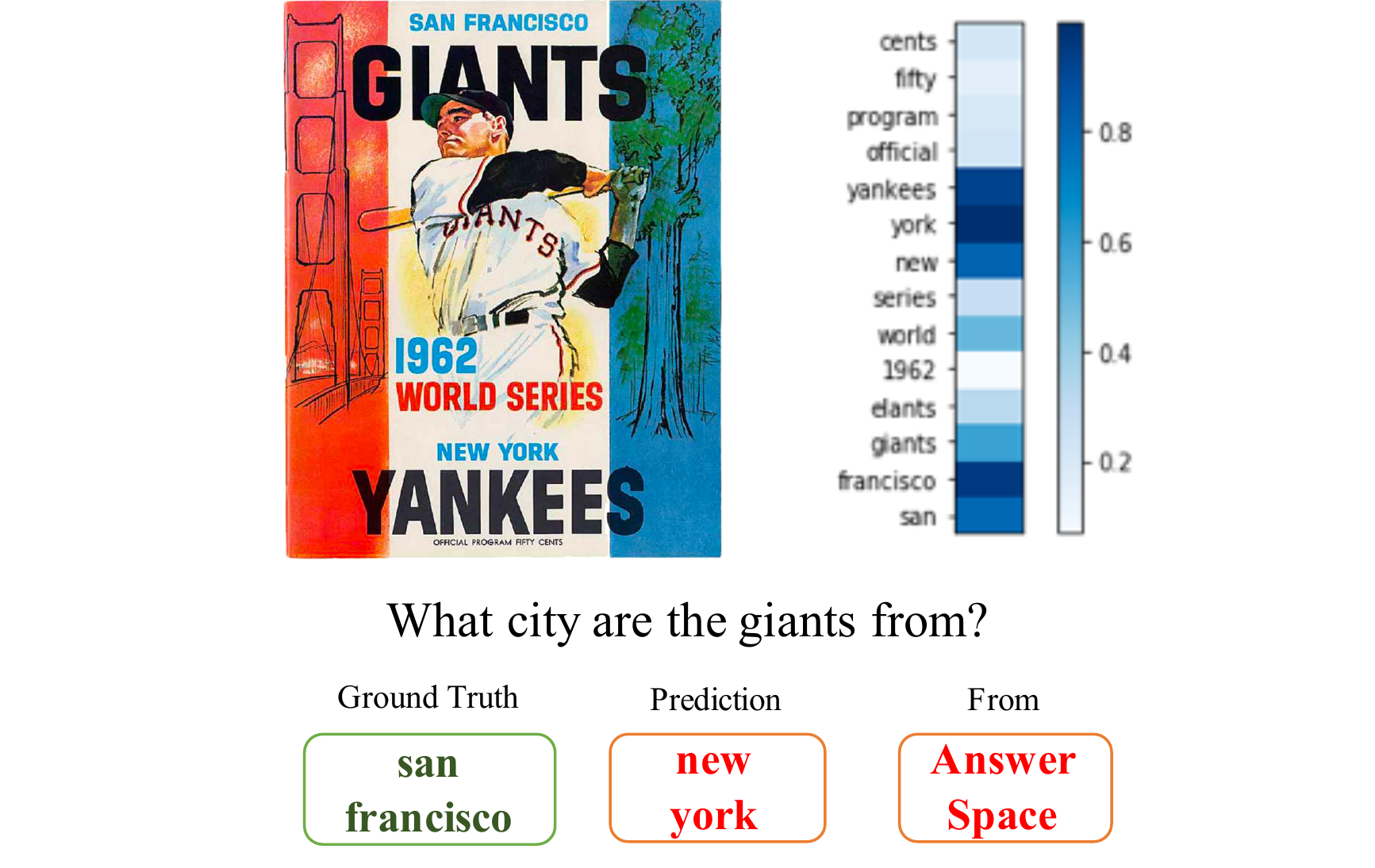}
        \caption{}
        \label{fig:attention_10}
    \end{subfigure}
    \begin{subfigure}[t]{0.32\textwidth}
        \includegraphics[width=1\linewidth]{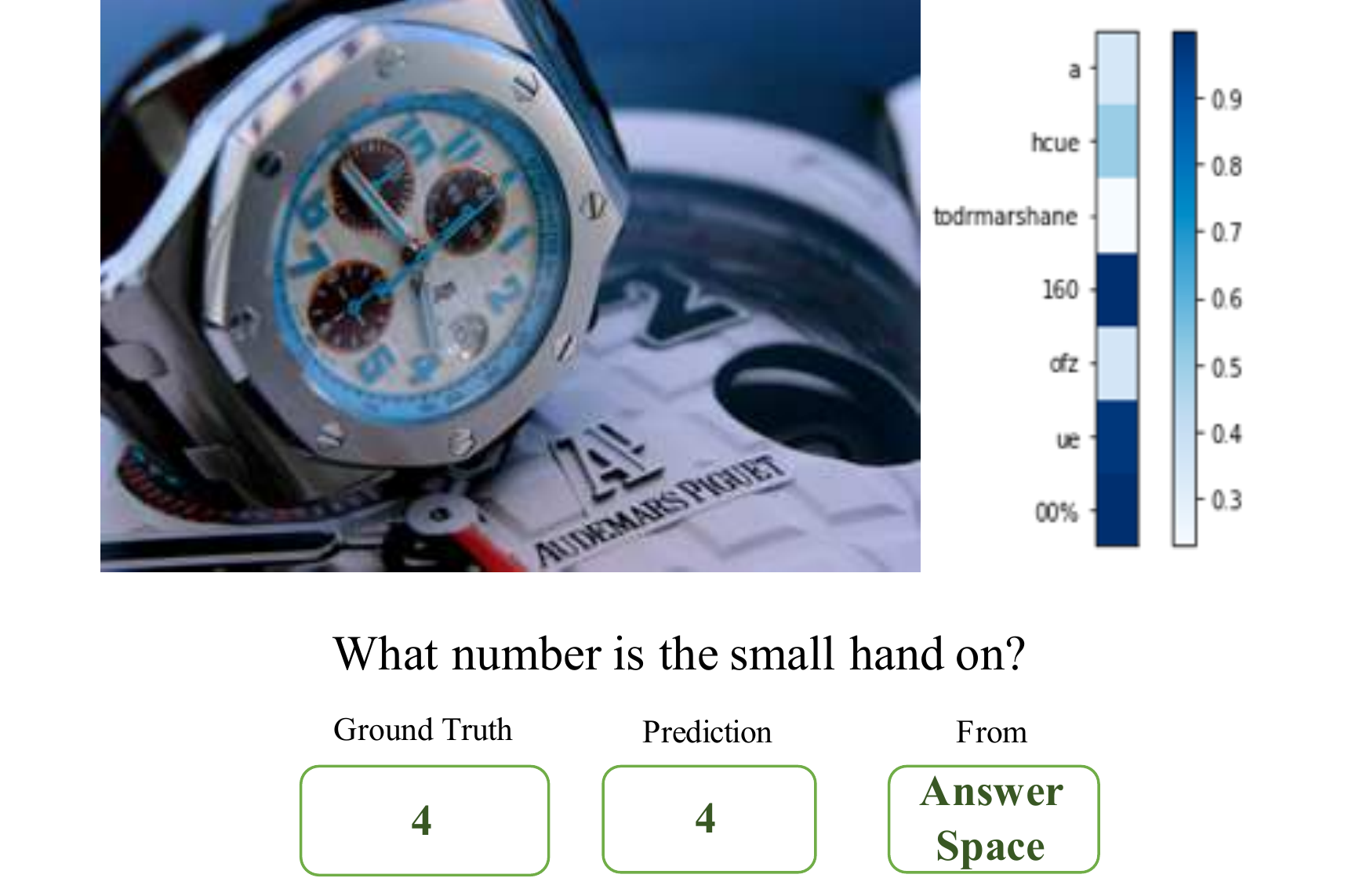}
        \caption{}
        \label{fig:attention_11}
    \end{subfigure}
    \begin{subfigure}[t]{0.32\textwidth}
        \includegraphics[width=1\linewidth]{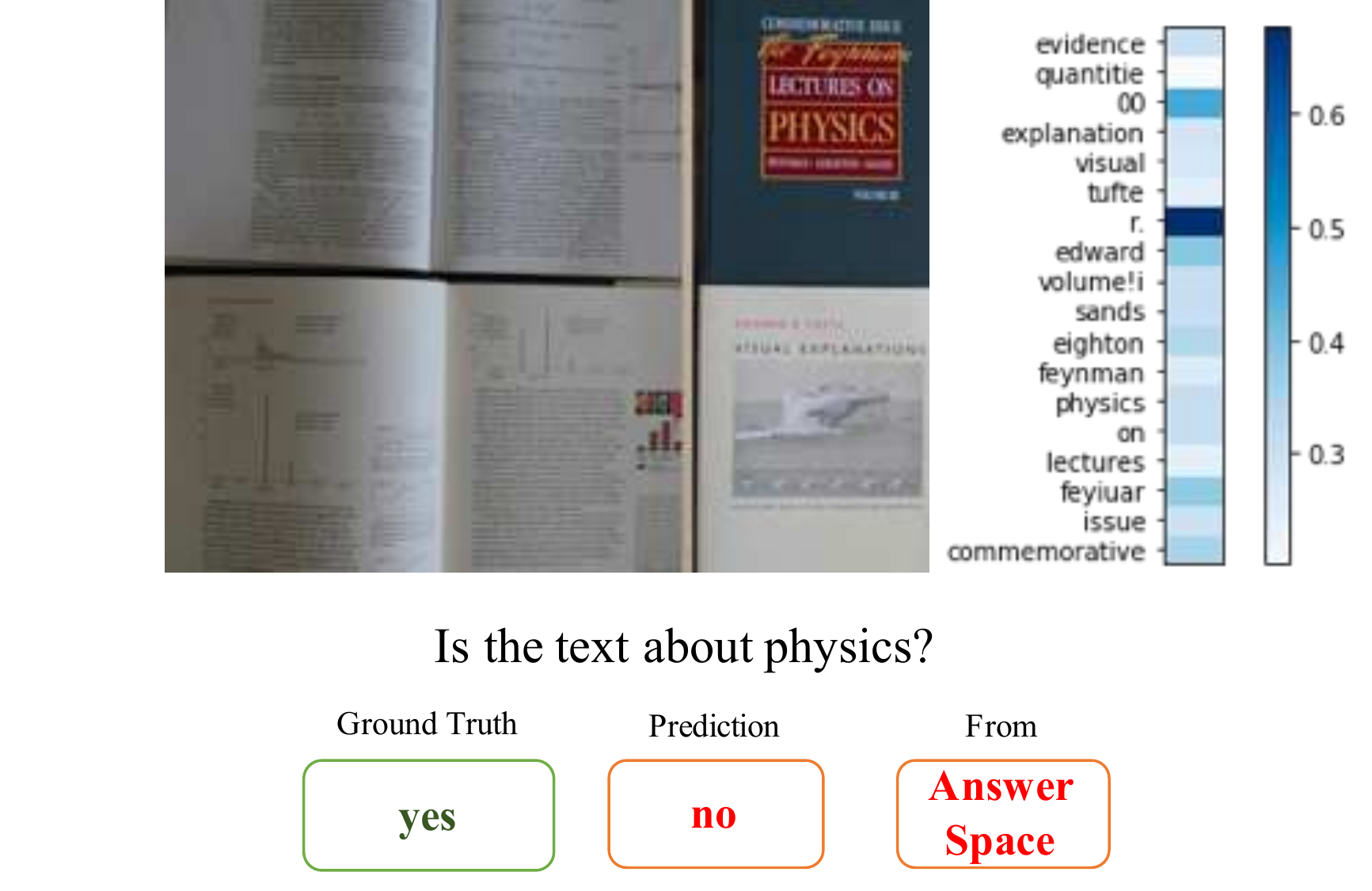}
        \caption{}
        \label{fig:attention_12}
    \end{subfigure}
    \begin{subfigure}[t]{0.32\textwidth}
        \includegraphics[width=1\linewidth]{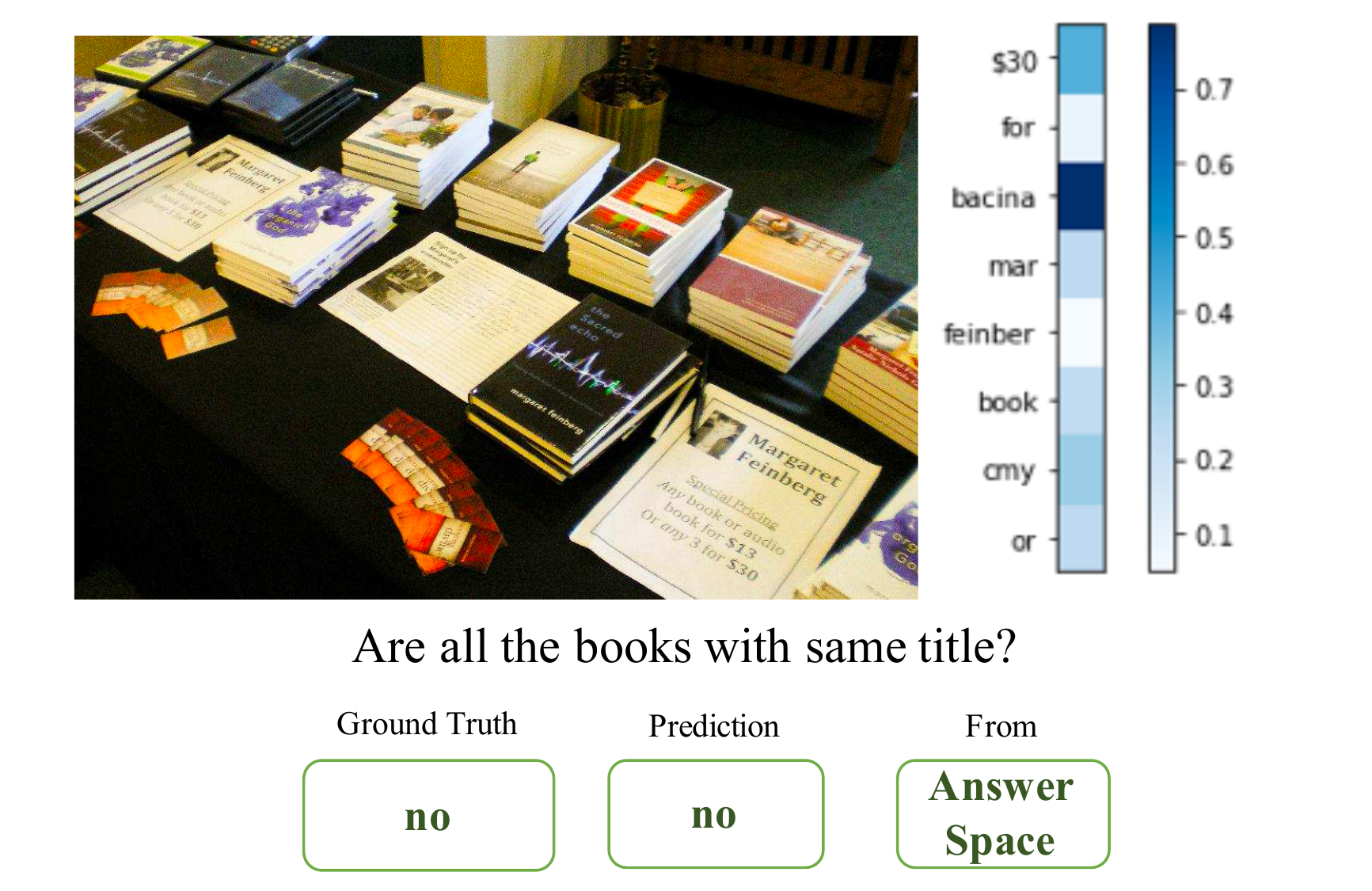}
        \caption{}
        \label{fig:attention_13}
    \end{subfigure}
    \begin{subfigure}[t]{0.32\textwidth}
        \includegraphics[width=1\linewidth]{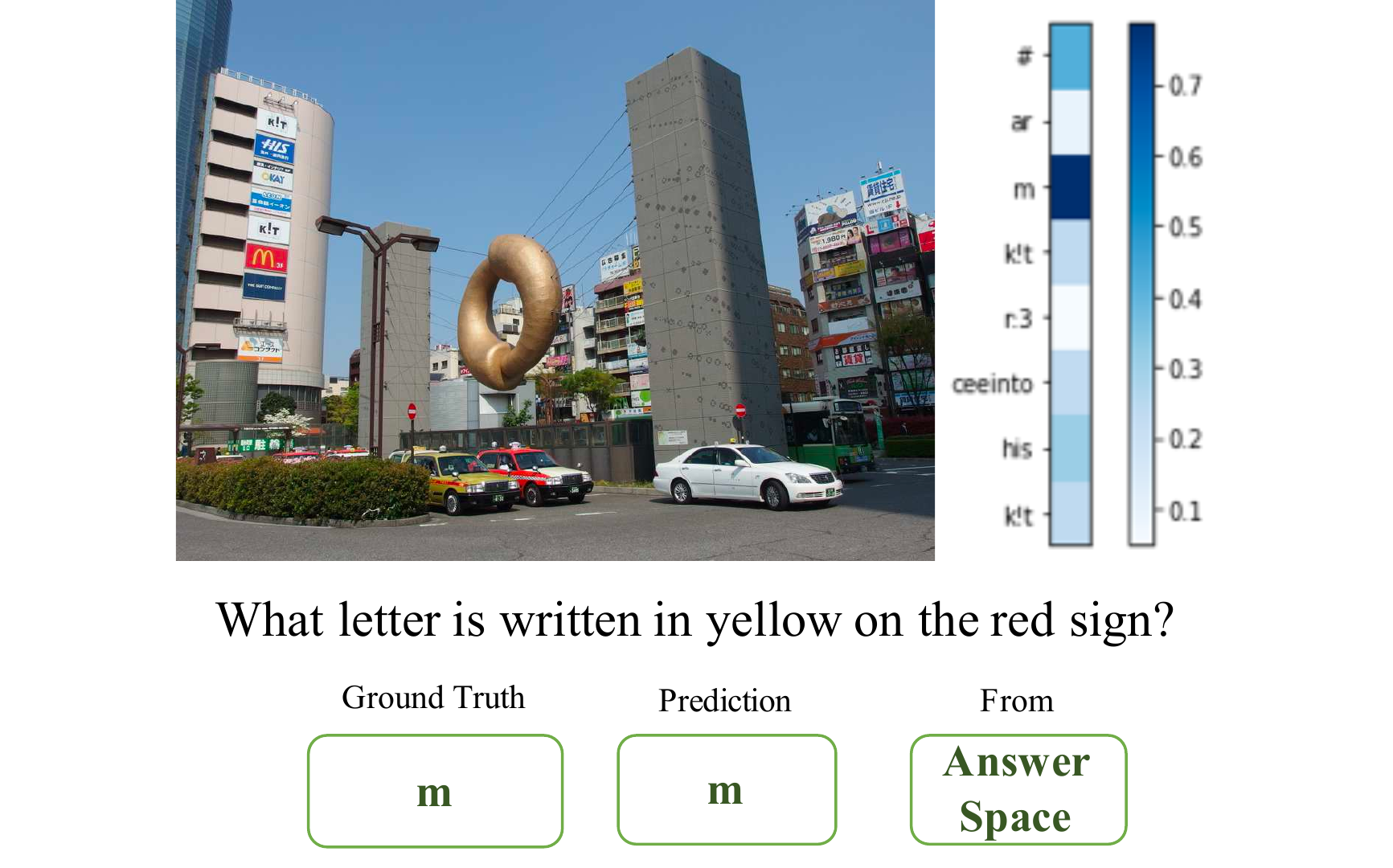}
        \caption{}
        \label{fig:attention_14}
    \end{subfigure}
    \begin{subfigure}[t]{0.32\textwidth}
        \includegraphics[width=1\linewidth]{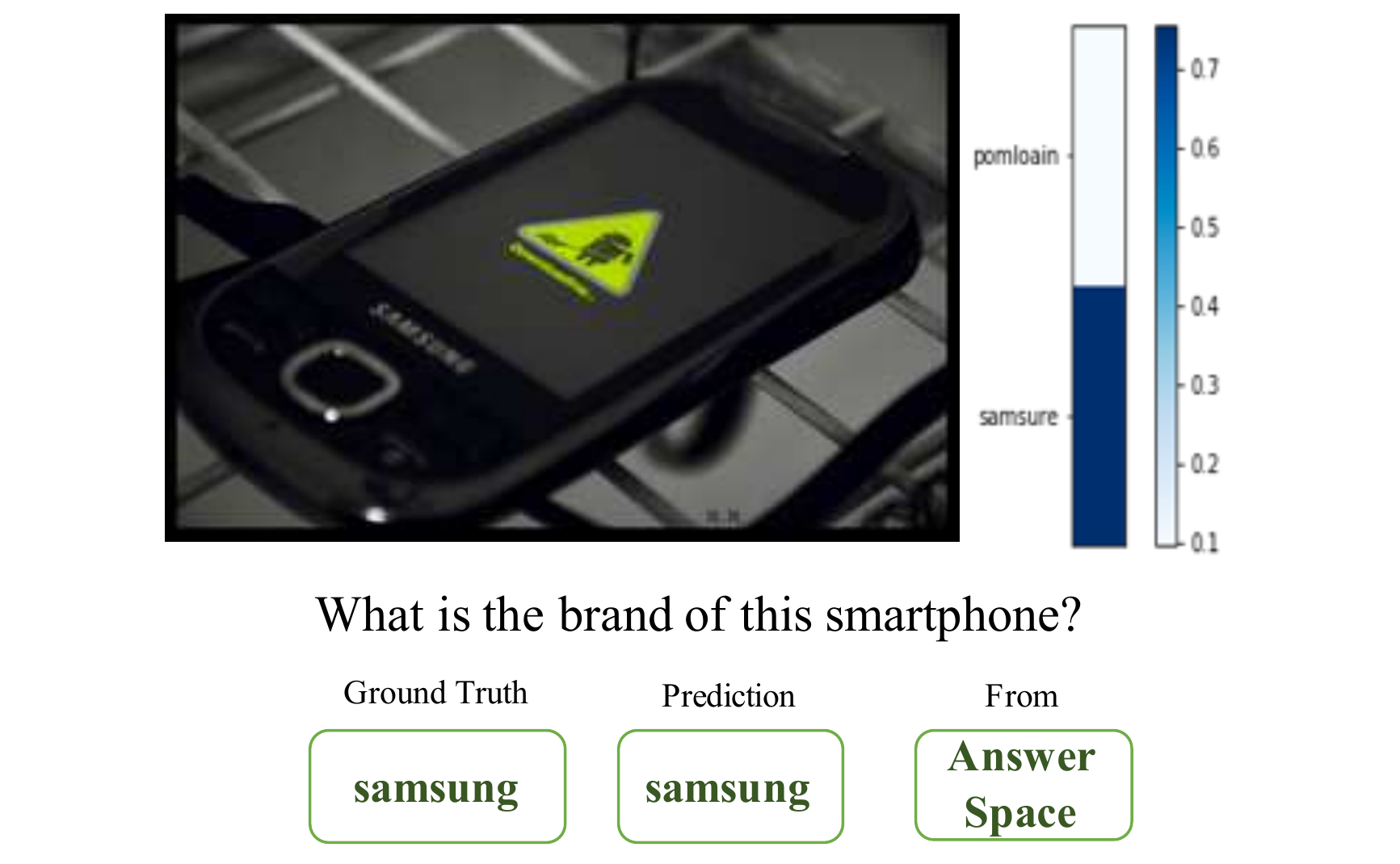}
        \caption{}
        \label{fig:attention_15}
    \end{subfigure}

    \label{fig:attention}
\end{figure*}
\begin{figure*}\ContinuedFloat
    \centering
    \begin{subfigure}[t]{0.32\textwidth}
        \includegraphics[width=1\linewidth]{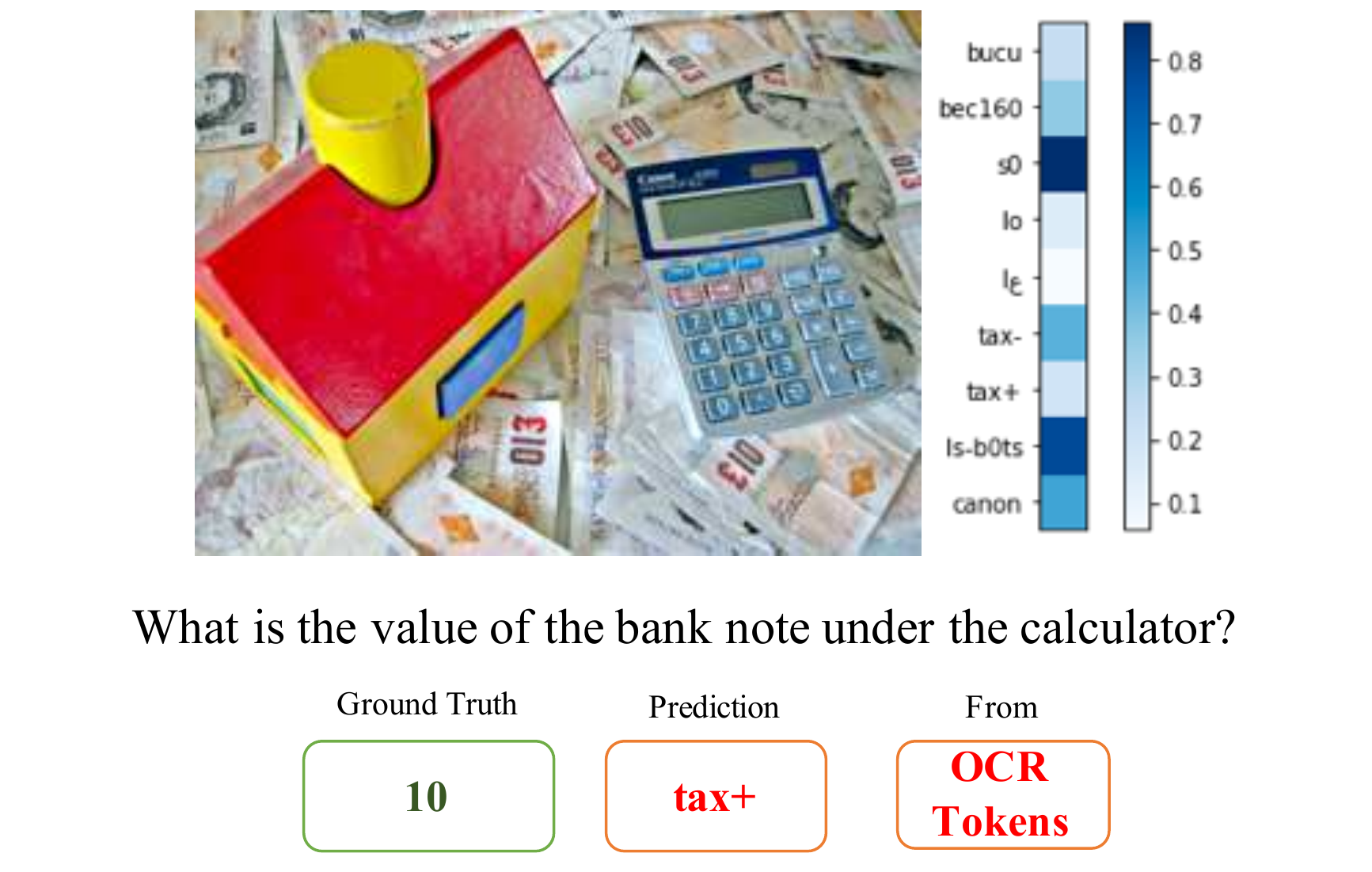}
        \caption{}
        \label{fig:attention_16}
    \end{subfigure}
    \begin{subfigure}[t]{0.32\textwidth}
        \includegraphics[width=1\linewidth]{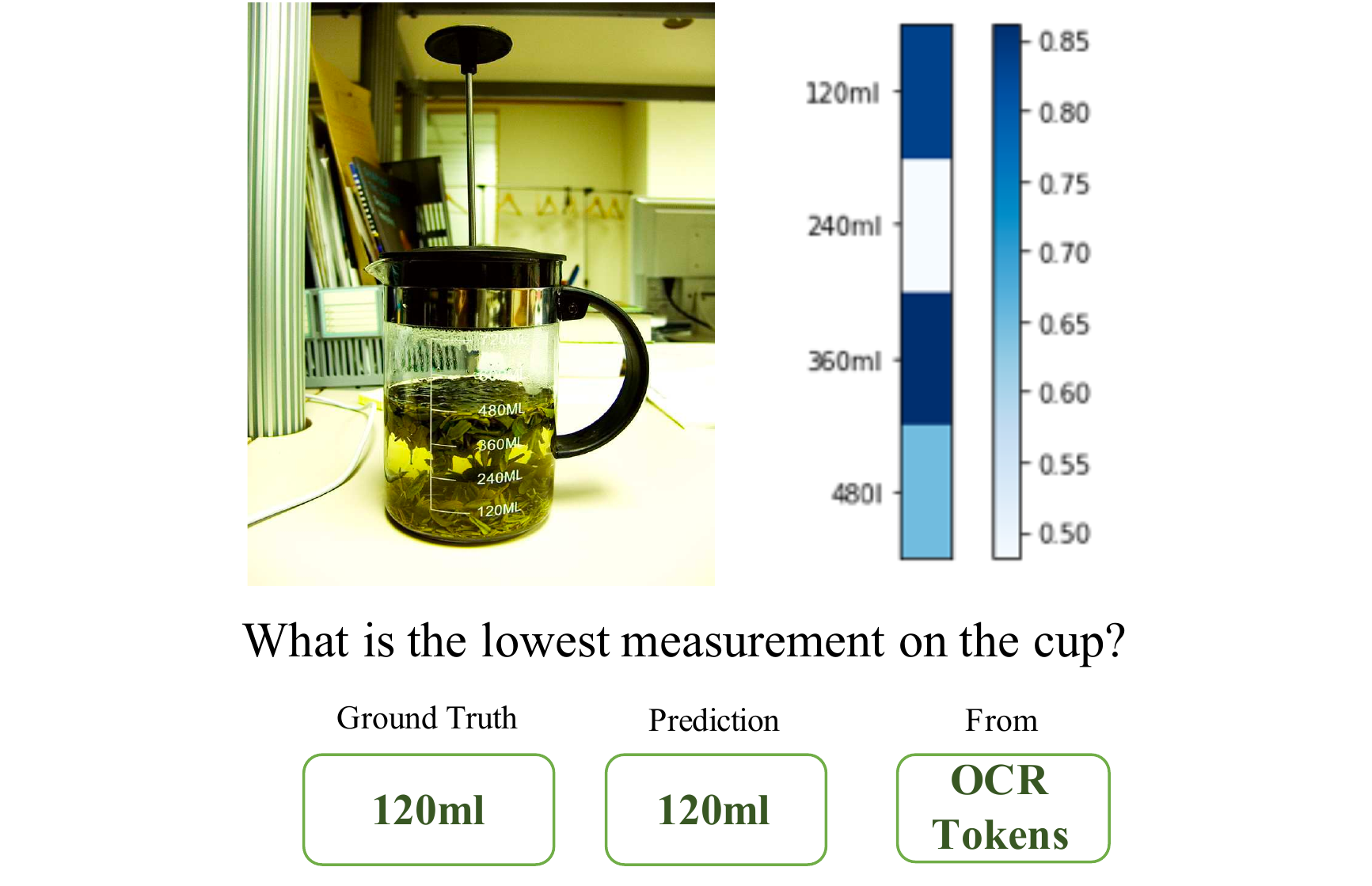}
        \caption{}
        \label{fig:attention_17}
    \end{subfigure}
    \begin{subfigure}[t]{0.32\textwidth}
        \includegraphics[width=1\linewidth]{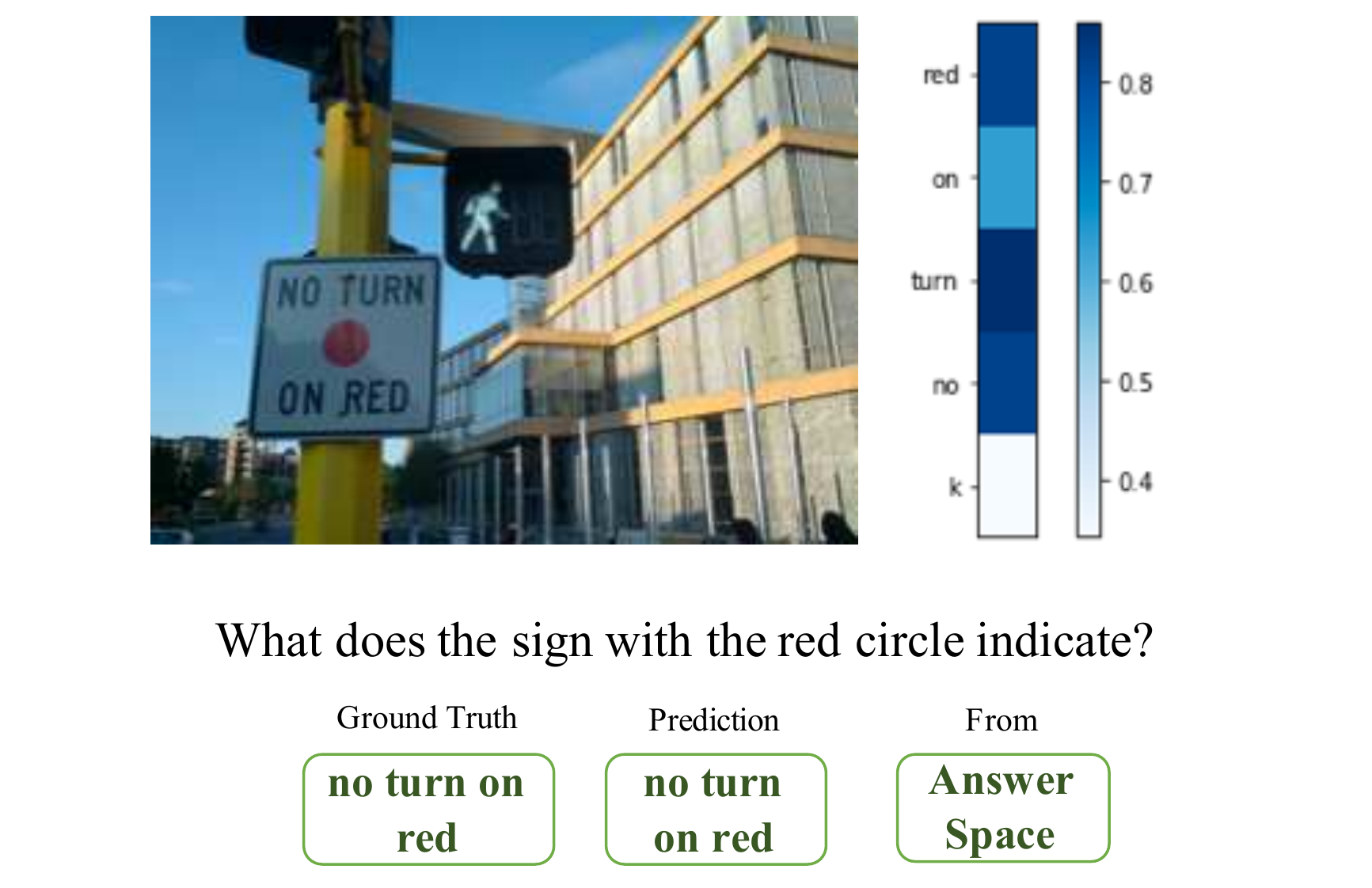}
        \caption{}
        \label{fig:attention_18}
    \end{subfigure}
    \begin{subfigure}[t]{0.32\textwidth}
        \includegraphics[width=1\linewidth]{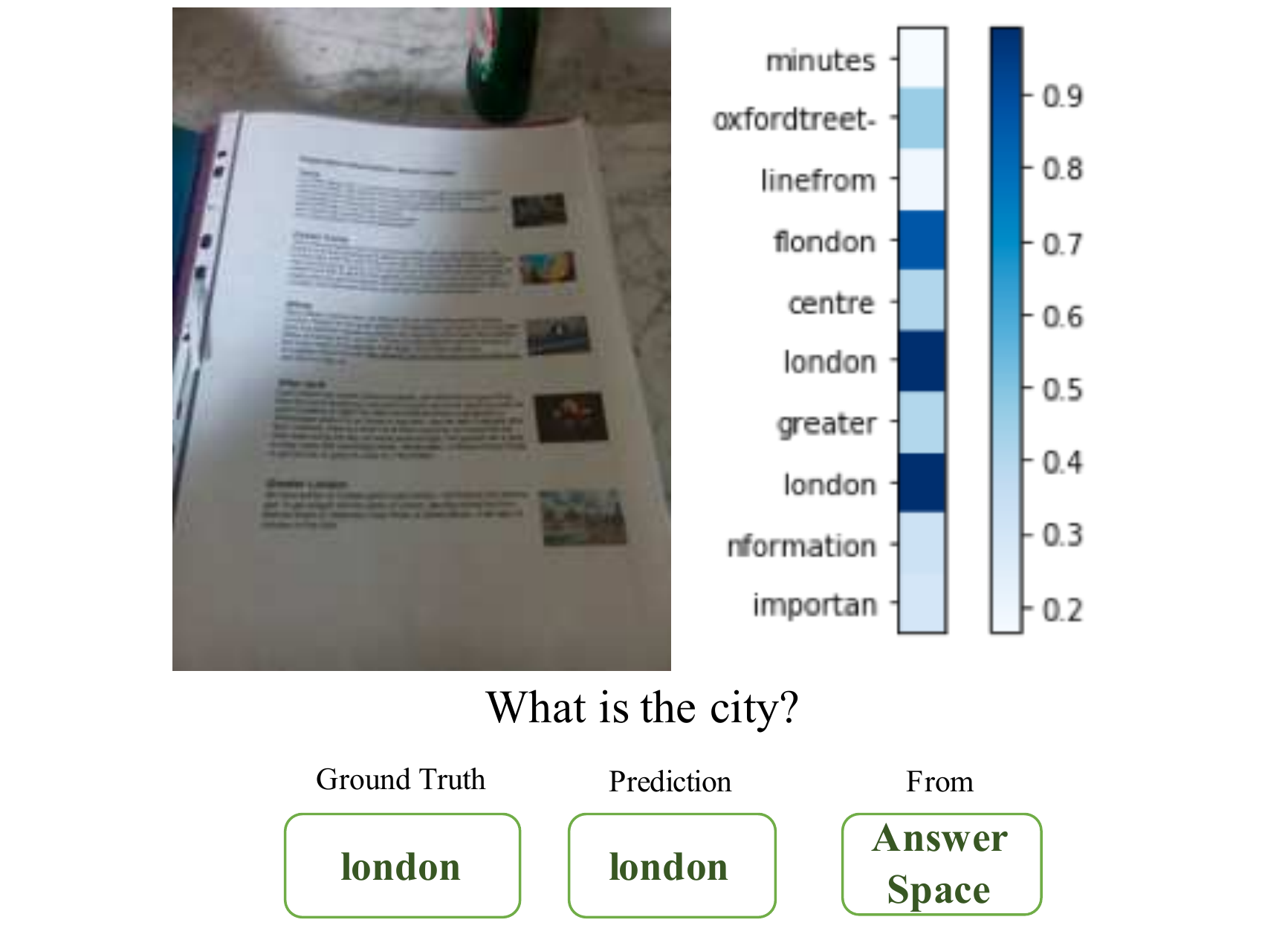}
        \caption{}
        \label{fig:attention_19}
    \end{subfigure}
    \begin{subfigure}[t]{0.32\textwidth}
        \includegraphics[width=1\linewidth]{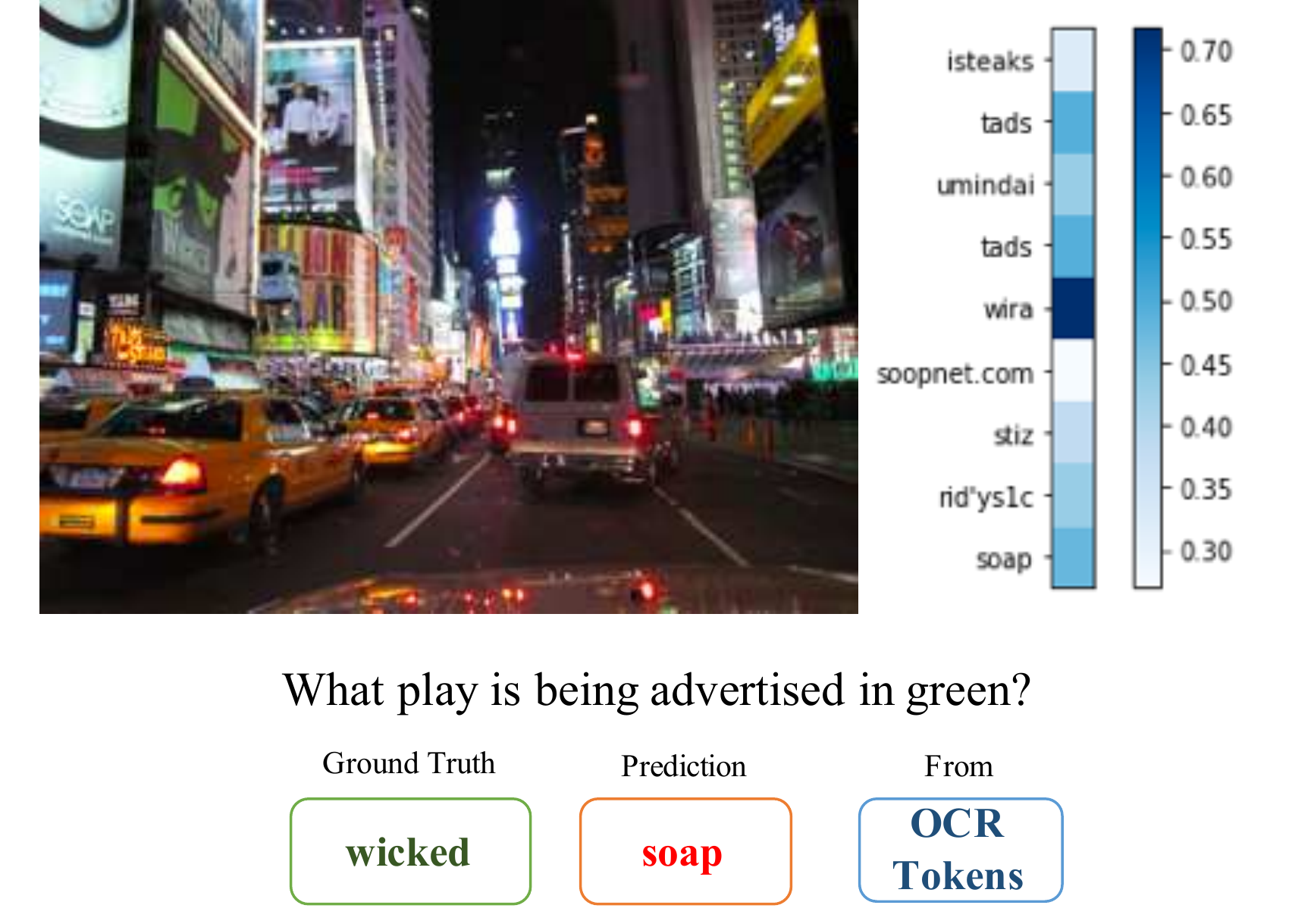}
        \caption{}
        \label{fig:attention_20}
    \end{subfigure}
    \begin{subfigure}[t]{0.32\textwidth}
        \includegraphics[width=1\linewidth]{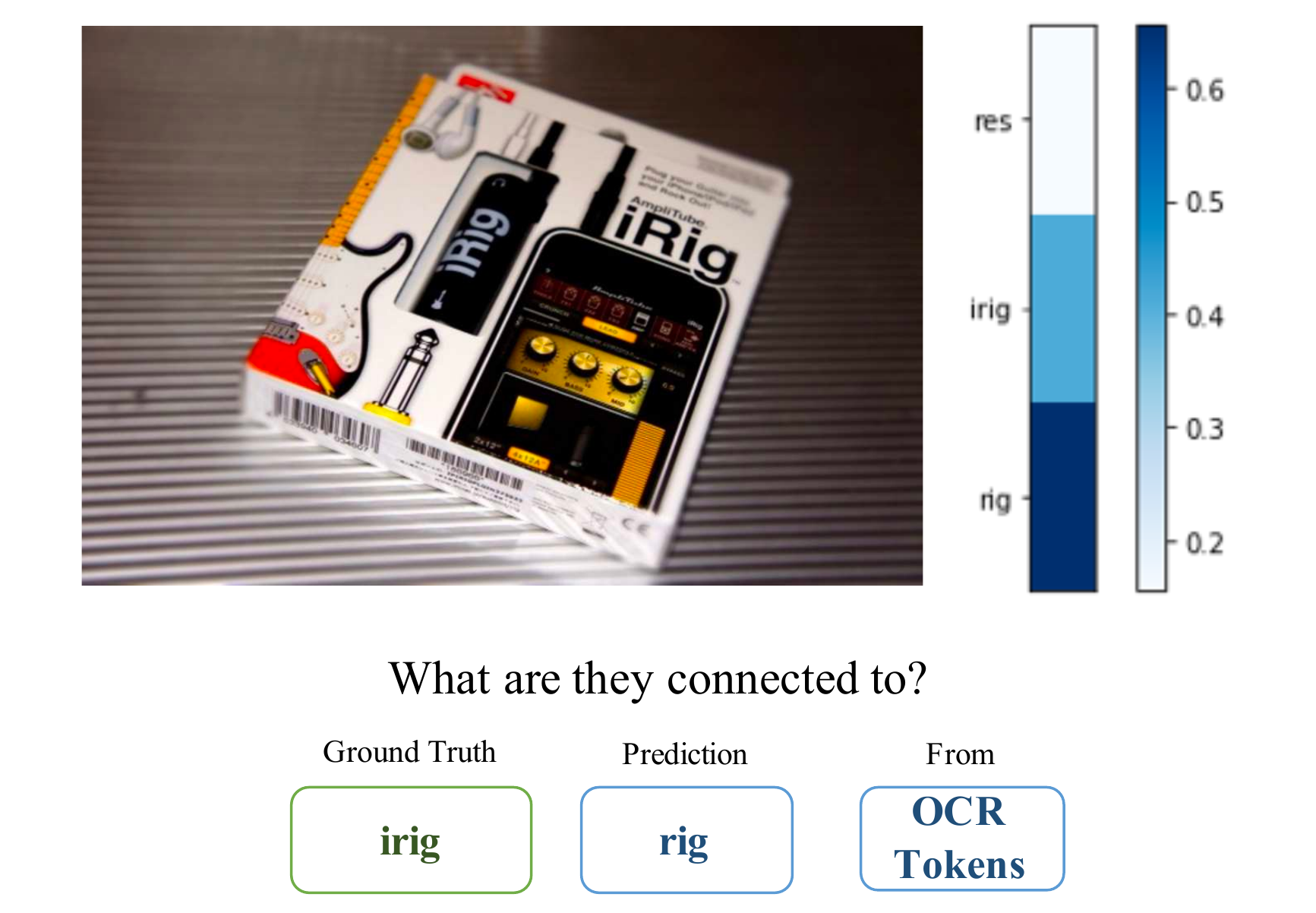}
        \caption{}
        \label{fig:attention_21}
    \end{subfigure}
    \caption{\textbf{\datasetName Examples and \approachNameShort's predictions on them.} We show multiple examples from \datasetName, ground truth answers, along with predictions from \approachNameShort, attention maps on OCR tokens and whether \approachNameShort predicted the answer from the OCR tokens or pre-determined answer space. 
    Green, red, and blue boxes correspond to correct, incorrect, and partially correct answers, respectively. On the right side of each image, we show attention bars which depict attention weights (0-1) for each of the OCR tokens.}
    
    \label{fig:attention_total}
\end{figure*}

\end{document}